\documentclass[10pt,twocolumn,letterpaper]{article}

\usepackage[pagenumbers]{cvpr} % To force page numbers, e.g. for an arXiv version

\usepackage{graphicx}
\usepackage{amsmath}
\usepackage{amssymb}
\usepackage{amsfonts}
\usepackage{booktabs}

\usepackage{mathtools}
\usepackage{xcolor}
\usepackage{pifont}
\usepackage{float}
\usepackage[accsupp]{axessibility}
\usepackage[pagebackref,breaklinks,colorlinks]{hyperref}

\usepackage[capitalize]{cleveref}
\crefname{section}{Sec.}{Secs.}
\Crefname{section}{Section}{Sections}
\Crefname{table}{Table}{Tables}
\crefname{table}{Tab.}{Tabs.}

\def\cvprPaperID{2472} % *** Enter the CVPR Paper ID here
\def\confName{CVPR}
\def\confYear{2023}

\begin{document}

%%%%%%%%% TITLE - PLEASE UPDATE

%\title{Scaling Up Learning on Graphs for  Multi-Object Tracking on Long Crowded Videos}
%\title{Towards Long-Term Tracking with Scalable and Unified Graph Hierarchies}
%\title{Unifying Short and Long-Term Tracking with Graph Hierarchies}
\title{Unifying Short and Long-Term Tracking with  Graph Hierarchies}

%\title{Unifying short and long-term tracking with graph hierarchies}

% \author{Orcun Cetintas\\
% Technical University of Munich \\
% {\tt\small orcun.cetintas@tum.de}
% % For a paper whose authors are all at the same institution,
% % omit the following lines up until the closing ``}''.
% % Additional authors and addresses can be added with ``\and'',
% % just like the second author.
% % To save space, use either the email address or home page, not both
% \and
% Guillem Brasó\\
% Technical University of Munich \\
% {\tt\small guillem.braso@in.tum.de}
% \and
% Laura Leal-Taix\'{e}\\
% Technical University of Munich \\
% {\tt\small leal.taixe@tum.de}
% }

\author{Orcun Cetintas\textsuperscript{1}{\textbf{\textsuperscript{*}}}
%Technical University of Munich \\
%{\tt\small orcun.cetintas@tum.de}
% For a paper whose authors are all at the same institution,
% omit the following lines up until the closing ``}''.
% Additional authors and addresses can be added with ``\and'',
% just like the second author.
% To save space, use either the email address or home page, not both
\quad
Guillem Brasó{\textsuperscript{1}\textsuperscript{2}\textbf{\textsuperscript{*}}}
%Technical University of Munich \\
%{\tt\small guillem.braso@in.tum.de}
%\and
\quad
Laura Leal-Taix\'{e}{\textsuperscript{1}\textbf{\textsuperscript{$\dagger$}}} 
\\
\textsuperscript{1}Technical University of Munich \quad \textsuperscript{2}Munich Center for Machine Learning
%name.surname@tum.de
%{\tt\small leal.taixe@tum.de}
}
%\maketitle

\newcommand{\lblfig}[1]{\label{fig:#1}}
\twocolumn[{%
\renewcommand\twocolumn[1][]{#1}%
\vspace{-3em}
\maketitle
\thispagestyle{empty}
\begin{center}
    \centering
    \includegraphics[width=\textwidth]{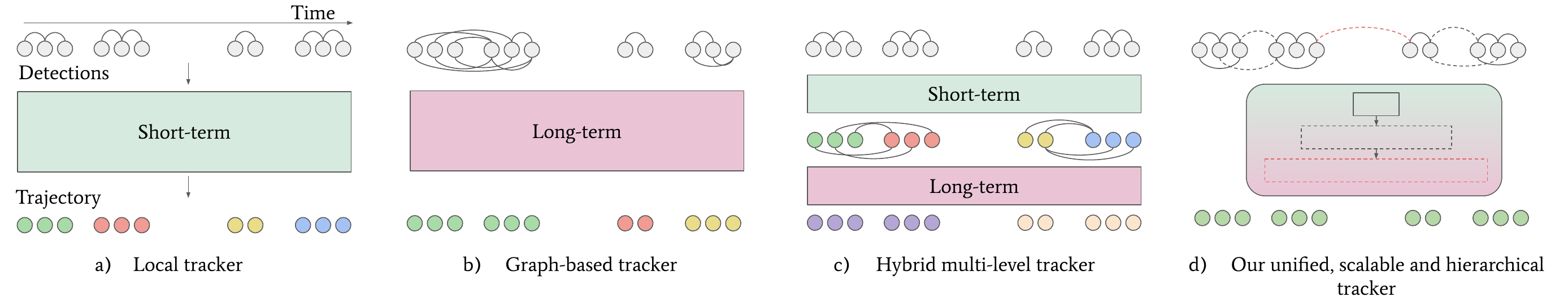}
    \vspace{-0.5cm}
    \captionof{figure}{ (a) Local tracker focusing on short-term scenarios and lacking robustness at long-term identity preservation (b) Graph-based tracker tackling longer-term association but unable to cover large time gaps due to its limited scalability  (c) Hybrid multi-level tracker engineering a combination of techniques but still struggling with scalability (d) Our unified hierarchical tracker with high scalability.} %to long-term association.}
    \lblfig{teaser}
    \label{teaser}
\end{center}%
}]

% Macros

\newcommand{\orc}[1]{{\leavevmode\color{teal}[Orcun: #1]}} % Orcun
\newcommand{\gui}[1]{{\leavevmode\color{blue}[Guillem: #1]}} % Guillem
\newcommand{\lau}[1]{{\leavevmode\color{magenta}[Khaleessi: #1]}} % Guillem
\newcommand{\old}[1]{{\leavevmode\color{gray}[Old text: #1]}} % Guillem
\newcommand{\REF}{\textcolor{red}{\textbf{REF}}}
\newcommand{\TODO}[1]{{\leavevmode\color{brown}[TODO: #1]}} % TODO
\newcommand{\PAR}[1]{{\noindent{\textbf{#1}}}} % TODO

\newcommand{\modelname}{SUSHI\@\xspace} % Our model's name
\newcommand{\modelnamenospace}{SUSHI} % Our model's name
\newcommand{\blockname}{SUSHI block\@\xspace} 
\newcommand{\blocknameplural}{SUSHI blocks\@\xspace} 

\newcommand{\cmark}{\ding{51}} % Checkmark
\newcommand{\xmark}{\ding{55}} % Cross
\newcommand{\eqdef}{\vcentcolon=}
\newcommand*{\mlp}{{\text{MLP}}}

%%%%%%%%% ABSTRACT
\begin{abstract}
Tracking  objects over long videos effectively means solving a spectrum of problems, from short-term association for un-occluded objects to long-term association for objects that are occluded and then reappear in the scene.
Methods tackling these two tasks are often disjoint and crafted for specific scenarios, and top-performing approaches are often a mix of techniques, which yields engineering-heavy solutions that lack generality. 
In this work, we question the need for hybrid approaches and introduce SUSHI, a unified and scalable multi-object tracker. Our approach processes long clips by splitting them into a hierarchy of subclips, which enables high scalability. 
We leverage graph neural networks to process all levels of the hierarchy, which makes our model unified across temporal scales and highly general.
As a result, we obtain significant improvements over state-of-the-art on four diverse datasets. Our code and models are available at \href{https://bit.ly/sushi-mot}{bit.ly/sushi-mot}. %, including \TODO{specific numbers}
\end{abstract}

\makeatletter{\renewcommand*{\@makefnmark}{}
\footnotetext{* Equal contribution.}\makeatother}

\makeatletter{\renewcommand*{\@makefnmark}{}
\footnotetext{$\dagger$ Currently at NVIDIA.}\makeatother}

%%%%%%%%% BODY TEXT

% Introduction
\vspace{-0.2cm}
\section{Introduction}
\label{sec:introduction}

Multi-Object Tracking (MOT) aims to identify the trajectories of all moving objects from a video. It is an essential task for many applications such as autonomous driving, robotics, and video analysis. Tracking-by-detection is a commonly used paradigm that divides the problem into (i) detecting objects at every frame and (ii) performing data association, i.e., linking objects into trajectories. 
%\textcolor{red}{Even if data association seems like one step that would require one solution, it actually contains two distinct problems that are often tackled with two distinct solutions: \textit{short-term} and \textit{long-term association}}. \orc{not the biggest fan of this sentence but couldn't come up with a better one.} \gui{I also don't like this sentence. I think that the claim is too strong/not correct: it's not true that data association contains \textbf{two} problems (we actually argue that it's a spectrum). Moreover, we later argue that some methods only tackle e.g. short term, so I don't think we can say that data association, in general, includes both problems. Does it make sense?} \TODO{proposal}
%

In the presence of highly accurate object detections, data association happens mostly among detections that are close in time, i.e., \textit{short-term association}. Simple cues such as position and motion-based proximity \cite{sort, tracktor, centertrack, Tokmakov_2021_ICCV, bytetrack} or local appearance \cite{jde, deepsort, zhang2021fairmot, qdtrack} are often enough for accurate association.
%
%
%
%Even appearance-based cues, e.g., obtained with re-identification methods \REF, work better for detections that are close in time, since appearance does not change drastically. \lau{I am writing this because we have QDtrack, right? I hate my sentence btw but it was just to put smthg there.} \gui{I don't think we need an entire sentence here to cover QDTrack, since it's a quite 'exotic' paper. Just added 'local appearance' before}
%
Different challenges appear in crowded scenes, when objects may be often occluded and not detected for several frames. This forces methods to perform association among detections in distant time frames, i.e., \textit{long-term association}. 
This requires specific solutions that build more robust global appearance models \cite{deepcc, Tang_2017_CVPR, deepmatching},  %\gui{not sure what to put here. Maybe stuff based on pose, densematching, etc.?}, 
create motion models capable of long-term trajectory prediction~\cite{Robicquet2016LearningSE, Leal-Taixe_2014_CVPR, quovadis_patrick} %\lau{cite some of patricks type of work}, 
or bring robustness by performing association across all frames and all trajectories using a graph representation~\cite{subgraph, mpntrack, lpc, gmt, Tang_2017_CVPR}. %\textcolor{red}{While more robust, graph-based approaches are generally limited by their scalability}. \lau{I would not talk about graphs until the limitations section. Here we are just presenting different types of trackers.}
Due to the different nature of these tasks, solutions used for \textit{short-term} associations tend to fail in \textit{long-term} scenarios. 
In fact, most state-of-the-art trackers use a combination of approaches to track over different timespans and therefore can be considered to be \textit{multi-level trackers}. 
%\orc{Why are we calling them multi-level and not hybrid etc.? Because our model has also "hierarchical levels". If multi-level refers to multiple time-scales we still need a name for these "combined methods" and this should be introduced in the intro}. %\lau{I have added the previous sentence back, I still feel we need it to link sentences.}
Several short-term trackers use independent re-identification (reID) mechanisms for long-term association \cite{tracktor, deepsort, Xu_2020_CVPR, Son_2017_CVPR, trackformer}.  Analogously, various graph approaches rely on local trackers to perform short-term association~\cite{mpntrack, lift, aplift}. %entirely \c or generate tracklets \REF as an initial step. 
All of these \textit{hybrid multi-level} approaches have two main limitations.

The first one is \textbf{scalability} since current methods cannot deal with long videos. %\gui{clips? We also don't scale to very long entire videos.}. \lau{how about just long videos? if we now say clips without introducing what it means, it sounds weird.}
%
%\textcolor{red}{As we increase the timespan between detections to be associated, the performance of reID-based methods degrades, since appearance can be extremely different between the two detections due to pose, camera, scale, or illumination changes. 
%
%Obtaining enough training data to cover all the aforementioned cases to obtain a truly robust reID model is a hard task, therefore, we consider current methods for reID not to be scalable to arbitrary timespans.} 
As we increase the timespan between detections to be linked, association becomes more ambiguous due to significant appearance changes and large displacements. Hence, local trackers using a handcrafted combination of appearance and motion cues %, and not relying on contextual information, 
will fail to scale to arbitrary timespans.
Graph-based methods are more robust, but association for large timespans entails the creation of very large graphs (even if combined with local methods), which is infeasible both computationally and memory-wise. %\orc{check grammar} %\gui{memory wise?} \gui{even when used in combination with local methods}

The second limitation is \noindent\textbf{generality}. Using different techniques for different timespans requires making strong assumptions about the cues needed at each temporal scale, which limits the applicability of these approaches. 
%
%For instance, in tracking scenarios where persons dress uniformly, such as dancing videos, motion is more reliable than appearance \REF, unlike videos with heavy camera motion, where appearance is the most reliable cue \REF. 
%
%
%
%
For instance, in tracking scenarios where people dress uniformly and frame rate is high, e.g. dancing videos~\cite{dancetrack}, proximity or motion-based local trackers \cite{tracktor,centertrack, bytetrack, sort} are more reliable than appearance-based trackers. 
On the other hand, in the presence of heavy camera motion or low frame rates, the performance of the aforementioned trackers degrades significantly, and appearance may become the most reliable cue~\cite{qdtrack, bdd}. 
%\orc{"most" id a bit daring, any refs?} cue \REF. 
%
Overall, these discrepancies lead inevitably to handcrafted solutions for each type of scenario.

In this work, we ask the following question: \textit{can we design a unified method that generalizes to multiple timespans and further scales to long videos?}

We propose a method that processes videos hierarchically: lower levels of our hierarchy focus on short-term association, and higher levels focus on increasingly long-term scenarios. 
The key differences to existing \textit{hybrid} multi-level solutions is that we use \textit{the same learnable model} for all time scales, i.e., hierarchy levels. 
Instead of handcrafting different models for different scales, we show that our model can learn to exploit the cues that are best suited for each time-scale in a data-driven manner.
Furthermore, our hierarchy allows a finer-grained transition from short to long-term instead of using two distinct stages. %\orc{Just added this sentence, does it make sense?}\gui{I like it!}. 
Our method targets the two main limitations of previous works: (i) its hierarchical structure makes it highly scalable and enables processing long clips efficiently, and (ii) it is highly general and does not make any assumptions about which cues are best suited for which timespans, but instead allows the model to obtain the necessary cues in a data-driven manner. 
%
%
%\lau{How about this:}
We, therefore, obtain a {\textbf{\underline{S}}trong tracker, with a {\textbf{\underline{U}}nified solution across timespans, and good \textbf{\underline{S}}calability thanks to its \textbf{\underline{HI}}erarchical nature, and name it \textbf{\modelname}.

At its core, \modelname is a graph method%\textcolor{red}{Current top-performing graph-based methods do not exploit the full potential of graph-based solutions since they use a single-monolithic graph \REF.
%We believe that the true power of graph-based solutions has not been unlocked yet because all works focus on the use of a single-monolithic graph. %\gui{Need to relax this sentence. We are not the first to propose a hierarchy}
%This has several limitations, mostly when trying to solve long occlusions. \textbf{Firstly}, it requires processing a large number of frames and detections, resulting in large graphs with a severe imbalance of correct and incorrect edge hypotheses. Such graphs are both memory-intensive and challenging to operate on for learning methods. 
%\textbf{Secondly}, using a single graph implies using the same cues both for short-term and long-term association, and a single graph needs to figure out which ones to use in which scenario.  %Furthermore, the cues used for short-term and long-term association are the same, and a single graph needs to figure out which ones to use in which scenario %\orc{I didn't understand this sentence}. \gui{rebuild sentence}
%Furthermore, the cues used for short-term and long-term association are the same, and a single graph needs to figure out which ones to use in which scenario \orc{I didn't understand this sentence}. \gui{rebuild sentence}
%
%We instead embrace the different nature of the different time scales data association cases, and exploit that by operating on a hierarchy of smaller graphs.}\gui{propose: 
, but instead of working on a single monolithic graph, we embrace the different nature of data association over different timespans and operate on a hierarchy of  graphs. At the lowest level of our hierarchy, nodes %in our graphs
represent object detections in nearby frames. We use a graph neural network (GNN) \cite{scarselli_gnn, gilmer2017neural, mpntrack} to process those into short tracklets, and then build new graphs to generate increasingly longer trajectories at every level of our hierarchy. Notably, we use \textit{the same GNN architecture at every level}. Thus we do not make any assumptions about what cues are best for each timespan.
%
%Overall, our method exploits the fundamental differences of tracking over different time scales to design a unified hierarchical model that can efficiently tackle long sequences. 
We demonstrate the generality of our approach by showing significant identity preservation improvements over the state-of-the-art in several highly diverse benchmarks: up to \textbf{+4.7} IDF1 on MOT17\cite{motchaijcv}, \textbf{+9.1} IDF1 on MOT20\cite{mot20}, \textbf{+9.5} IDF1 on DanceTrack\cite{dancetrack}, and \textbf{+4.2} IDF1 on BDD\cite{bdd}, therefore setting new state-of-the-art results by a significant margin.  % We will release our code and models. %\TODO{specific numbers}

\begin{figure*}[ht]
\centering
\includegraphics[width=\textwidth]{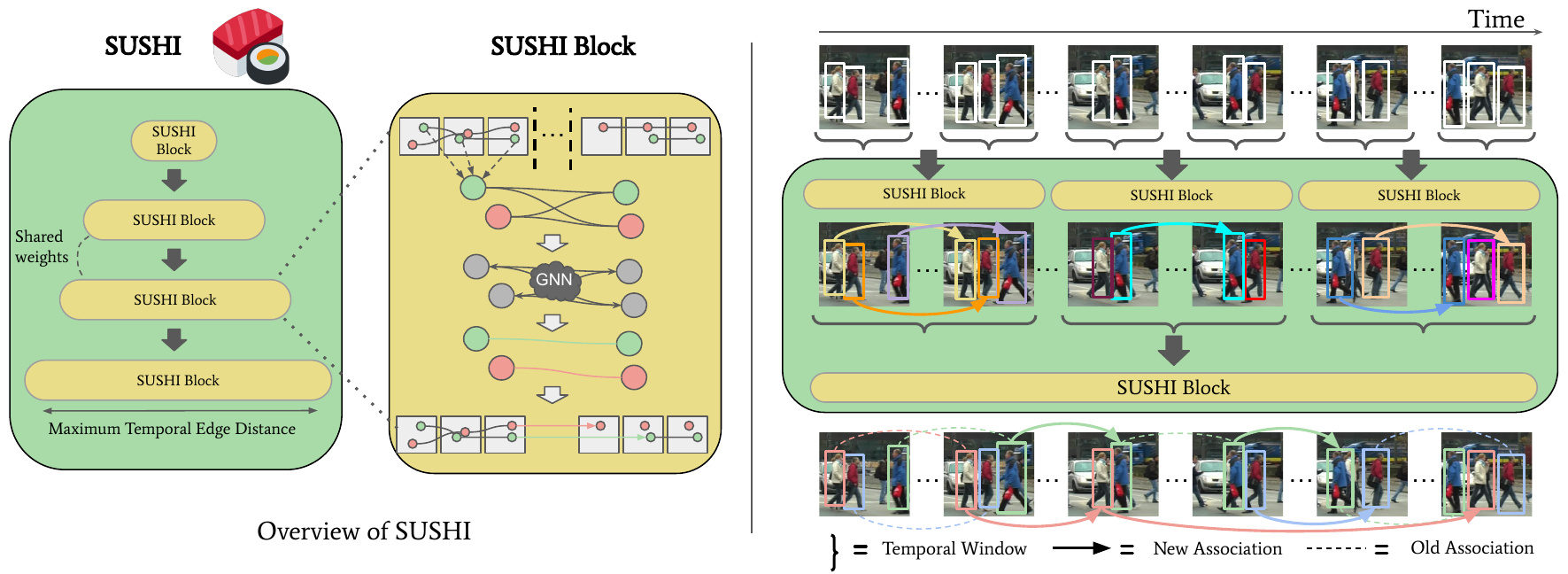}
\caption{ %SUSHI consists of SUSHI blocks each corresponding to a hierarchy level. SUSHI blocks share the same weights and maximum temporal edge distance within the graphs increases at each level. A sushi block considers a graph with tracklets as nodes, performs message
%passing and links the nodes belonging to the same identity. Overall, by hierarchically stacking several SUSHI blocks we progressively merge detections and tracklets into full trajectories
\modelname consists of a set of \blocknameplural operating hierarchically over a set of tracklets (with initial length one) in a video clip. Each \blockname considers a graph with tracklets from a subclip as nodes, performs neural message passing over it, and merges nodes into longer tracks. Over several hierarchy levels \blocknameplural are able to progressively merge tracklets into tracks spanning over the entire clip. Notably, \textit{\blocknameplural share the same GNN architecture and weights}, hence making \modelname unified across temporal scales. }
\label{fig:pipeline}
\end{figure*}
\section{Related Work}

%Tracking-by-detection is a common paradigm in MOT that, given an input video, divides the task into: (i) detecting objects at each frame~\cite{fasterrcnn, yolo, detr}, and (ii) linking them into temporally coherent trajectories. 
%
%Most MOT work focuses on the second task, namely, data association.

\PAR{Short-term tracking.} Numerous modern trackers use frame-by-frame online association frameworks ~\cite{sort, tracktor, centertrack, Tokmakov_2021_ICCV, bytetrack, qdtrack, jde, zhang2021fairmot, trackformer, motr, Xu_2019_ICCV}. Motion and spatial proximity cues tend to be central components of these trackers. Notable examples include the widespread use of kalman-filter based motion models~\cite{sort, bytetrack, zhang2021fairmot, jde} or frame-by-frame regression-based frameworks ~\cite{tracktor, centertrack, trackformer, motr}. Some trackers further rely on appearance to increase robustness at lower frame-rates or under strong camera movement ~\cite{qdtrack, jde, zhang2021fairmot, Xu_2019_ICCV, deepsort}. While having good performance in short-term scenarios, these trackers lack robustness when it comes to long-term identity preservation.

\PAR{Graph-based tracking.} Graphs are a commonly used framework to model data association. They model nodes as object detections and edges as trajectory hypotheses. In contrast to frame-by-frame trackers, graph-based methods search for \textit{global} solutions to the data association problem over several frames and are therefore more robust.
%~\cite{global_traj_optim, berclaz2011multiple, network_flows_tracking, berclaz2006robust, 4270205, 5459278, zamir2012gmcp, subgraph}.
%Nodes represent object detections at different time frames, and edges model pairwise association hypotheses among them. 
%This formulation has been extensively studied to model the overall task as a discrete optimization problem~\cite{global_traj_optim, berclaz2011multiple, network_flows_tracking, berclaz2006robust, 4270205, 5459278, zamir2012gmcp, subgraph}. 
%More specifically, given a set of edge costs encoding the likelihood of each pair of objects being in the same trajectory, the overall task can be modelled as an edge labelling task. Following this view, several works have modelled the task as instances of optimization problems 
 To this end, numerous optimization frameworks have been studied, including network flows ~\cite{berclaz2011multiple, network_flows_tracking},  multi-cuts~\cite{Tang_2017_CVPR}, minimum cliques ~\cite{zamir2012gmcp}, disjoint paths~\cite{subgraph, lift, aplift}, and efficient solvers~\cite{berclaz2011multiple, Butt2013} have been designed. %\textcolor{red}{Recently, \REF proposed a new formulation: the lifted disjoint paths problems, as well as an approximate solver for it \REF}. 
In our work, we rely on a simplified version of the min-cost flow formulation~\cite{network_flows_tracking, mpntrack}, which allows us to avoid expensive optimization and use a small-scale linear program while still taking  advantage of graph-based tracking. %\gui{should mention scalability?} \lau{you mention it below no? i think it is not needed to mention twice. i would put more limitation fire in GNN}\gui{Wanna mention Global transformers somewhere!!}\lau{why?}

\noindent{\bf Learning in graph-based tracking.} While early graph-based methods focused on obtaining pairwise association costs from learning methods such as conditional random fields~\cite{crf_learned}, or handcrafted models~\cite{color_tracking}, recent approaches focus almost exclusively on deep learning techniques. Notable examples include learning pairwise appearance costs with convolutional networks ~\cite{Leal-Taixe_2016_CVPR_Workshops, Son_2017_CVPR, Ristani_2018_CVPR}, or learning track management policies with recurrent models \cite{Sadeghian_2017_ICCV, tracking_rnn}. Recently, numerous approaches learn models that natively operate on the graph domain such as graph neural networks (GNNs) ~\cite{mpntrack, lpc, gmt, graph_nets_mot, gnn_3dmot, gsm, braso2022multi} or transformers~\cite{zhou2022global}. %Among these works, MPNTrack~\cite{mpntrack} was the first to use GNNs on a graph of detections to be able to classify edges with high accuracy for MOT. 
While showing promise, current  %\lau{are they all the graph approaches or only the GNN approaches?} 
GNN and transformer-based  works have an important limitation: they operate over large monolithic graphs of detections, and therefore lack the scalability needed to process long video clips.

%and we adopt their GNN due to its simplicity and performance. Overall, while promising, all previous GNN-based works operated over large monolithic graphs of detections, and therefore lack of scalability prevented their applicability to long video clips.

%, our work differs in two key aspects: (i) MPNTrack operates on a single graph, while we propose to operate on a hierarchy of graphs, covering significantly longer video clips, and (ii) we propose a more general framework in which nodes are not restricted to be detections, but can instead be  tracklets of arbitrary length. Hence, we can exploit track-level cues such as motion, that were not available in~\cite{mpntrack}. 

%We are inspired by its work, and adopt its message passing framework for our tracking graphs. T
%While we adopt the same message passing framework, our work differs in two key aspects: (i) MPNTrack operates on a single graph, while we propose to operate on a hierarchy of graphs, covering significantly longer video clips, and (ii) we propose a more general framework in which nodes are not restricted to be detections, but can instead be  tracklets of arbitrary length. Hence, we can exploit track-level cues such as motion, that were not available in~\cite{mpntrack}. 

\noindent{\bf Multi-level hybrid tracking approaches.} %\gui{more precise bashing of old hierarchical methods} 
Multi-level tracking methods are dominated by handcrafted combinations of approaches. Several early tracking works exploited the idea of building tracks hierarchically  \cite{hierarchical_first, multi_view_hierarchical, hierarchical_hypergraph, net_flows_stitching}. They generally did so by generating short tracklets with handcrafted methods and then merging those within multiple stages involving different optimization techniques~\cite{hierarchical_hypergraph,multi_view_hierarchical,net_flows_stitching} and association cues~\cite{hierarchical_first}.  %so by using different optimization schemes ~\cite{hierarchical_hypergraph} and feature sources~\cite{hierarchical_first, multi_view_hierarchical}
%Several classical works use heterogeneous handcrafted techniques across different levels of hierarchical frameworks for data association \cite{hierarchical_first, multi_view_hierarchical, hierarchical_hypergraph}. 
In a similar fashion, numerous modern trackers combine several techniques to build tracks in a hierarchical, incremental way, without necessarily relying on graphs ~\cite{tracktor, trackformer, mpntrack, lpc, lift, aplift, associating_clips}. Some examples include short-term trackers using external networks for reID-based association \cite{deepsort, tracktor, trackformer}, and graph-based methods relying on local trackers to either generate tracklets ~\cite{mpntrack} or fill gaps in trajectories ~\cite{aplift, lift}. In our work, we show that we do not need to manually combine techniques, and instead, we use a unified GNN-based framework to efficiently perform data association across multiple hierarchical levels. %\gui{tracking by associating clips} 

%Motivated by the shortcomings of pre-deep learning object detectors, \cite{hierarchical_first} first proposed a framework to hierarchically merge object detections into three levels of increasingly long-term tracklets. It did so with a maximum a posteriori inference framework based on the seminal work of \cite{network_flows_tracking}, using the Hungarian algorithm. Building on the same Bayesian framework, \cite{multi_view_hierarchical} proposed a hierarchical approach for trajectory generation in the context of multi-camera tracking. Lastly \cite{hierarchical_hypergraph}  proposed to obtain long-term trajectories in single-camera tracking with a hierarchy of tracklets. They did so by constructing a graph with additional high-order edges, \ie a \textit{hypergraph}, and proposing a search-based discrete optimization algorithm for merging tracks based on hand-crafted features. While sharing a similar motivation to our approach, their graph construction, hierarchy, and overall model approach are significantly different from ours: we choose a simple hierarchy based on a recursive partition of video clips, and focus on learning how to merge trajectories, while relying on a very simple inference optimization algorithm.

% Methodology

\section{Background} \label{background}
\PAR{Tracking by detection.} Our approach follows the \textit{tracking-by-detection} paradigm. That is, given a video clip, we assume that object detections are computed for every frame, and our task is to perform data association by linking object detections into trajectories.
We denote the set of object detections as $\mathcal{O}$. Each object detection $o_i\in \mathcal{O}$ can be identified by its bounding box coordinates, its corresponding image region, and timestamp. %$(b_i, a_i, t_i)$.
%
%Its bounding box is given by its center coordinates, its width, and its height $b_i \eqdef (x_i, y_i, w_i, h_i)\in \mathbb{R}^4$.
%
% Its bounding box is given by its center coordinates, its width, and its height $b_i \eqdef (x_i, y_i, w_i, h_i)\in \mathbb{R}^4$.
%
The goal of the data association step is to obtain the set of trajectories $\mathcal{T}^*$ that group detections corresponding to the same identity. Each trajectory $T_k$ is given by its set of detections $T_k\eqdef\{o_{k_i}\}_{i=1}^{n_{k}}$, with $n_k$ being the number of object detections, or trajectory length. 

\PAR{Graph-based tracking. } Our method builds on the commonly used graph-based formulation of~\cite{network_flows_tracking}, which we briefly review.  
We model data association with an undirected graph $G=(V, E)$ in which each node corresponds to an object detection, \ie, $V \eqdef \mathcal{O}$. Edges represent association hypotheses among objects at different frames $E\subset\{(o_i, o_j)\in V\times V | t_i \neq t_j\}$. 
Formally, a time-ordered track $T_k=\{o_{k_i}\}_{i=1}^{n_k}$ with $t_{k_i}< t_{k_{i+1}}$, can be represented as a path in $G$ given by its edges $E(T_k) \eqdef\{(o_{k_1}, o_{k_2}), \dots, (o_{k_{n_k - 1}}, o_{k_{n_k}}) \}$. 
Therefore, edges $(u, v)\in E$ can be classified into correct hypotheses if $(u, v) \in E(T_k)$ for some $T_k\in \mathcal{T}^*$, in which case we denote $y_{(u, v)}=1$; or incorrect otherwise, \ie  $y_{(u, v)} = 0$. 
Given a set of edge predictions or \textit{costs}, $\{y^{\text{pred}}_{(u, v)}\}_{(u, v)\in E}$ aiming to estimate the set of edge labels $\{y_{(u, v)}\}_{(u, v)\in E}$, final trajectories can be obtained by rounding predictions into binary decisions via discrete optimization or heuristics, and then extracting their corresponding paths in the graph. Overall, this formulation allows casting data association as edge classification. 
%\lau{I did not check the notation in detail, its pretty unreadable in latex :D}

\section{\modelname}
% \emoji{sushi} 
\label{sec:method}

%\subsection{Method overview}
%\TODO{Tune/Re-write emphasizing selling points}
We provide an overview of \modelname in Fig. \ref{fig:pipeline}. It consists of a sequence of jointly trained \textit{\blocknameplural} that operate over a video clip.
Starting from initial per-frame object detections (referred to as tracklets of length one from now on) each \blockname learns to merge tracklets from the previous level into longer ones. 
To do so, each \blockname builds a graph in which the nodes represent tracklets from the previous level and the edges model trajectory hypotheses. 
Nodes and edges have associated embeddings encoding position, appearance, and motion cues that are propagated across the graph with a GNN. 
After several message-passing steps, edge embeddings are classified into correct and incorrect hypotheses, yielding a new set of longer tracklets. By hierarchically stacking several \blocknameplural, tracklets grow progressively into our final output: final tracks spanning over the entire  input video clip.

Overall, the GNN in each \blockname learns to exploit timespan dependent association cues and, combined, \blocknameplural enable tracking over long-time horizons efficiently.

\subsection{Constructing a hierarchy of tracking graphs} \label{section:graph_hierarchy}

%\PAR{The limitations of monolithic tracking graphs}. 
%Given an input video clip with $C$ frames, our goal is to be able to associate objects even if they are occluded for long time spans (i.e. up to $C-1$ frames). Given the graph formulation we introduced in \REF, this is only possible if we consider edges in our graph spanning across those all possible time distances (i.e. 1, 2, \dots, $C-1$). Doing so naively is prohibitively expensive for long sequences. Moreover, it implies that most edges will represent incorrect hypotheses, which will yield a severe edge label imbalance for learning methods. To prove that, it is enough to see that each object detection can, at most, be incident to two correct edges (one in the past, and one in the future). Therefore, in a clip with $\mathcal{O}$ object detections, the correct number of hypotheses is at most $2|\mathcal{O}|$, while the number of edges can grow until  $2|\mathcal{O}|$

%\gui{Maaybe separate paragraph explaining this:} Motivated by these problems, we instead propose building a hierarchy of smaller graphs to operate over long video clips. 

%\PAR{Hierarchical clip partitioning}. 
% Motivation for hierarchy.
\PAR{On the limitations of monolithic tracking graphs}. 
%A central problem in MOT is occlusions, where detections are not available for several frames, and we need to perform an association between the last frame where the object was seen and the frame where it reappears after the occlusion.
%To solve occlusions, given an input video clip with $C$ frames, our goal is to be able to associate object detections whose time distance spans as many frames as possible (i.e. up to $C-1$ frames). 
Given an input video clip with $C$ frames, our goal is to enable the association of objects even if they are occluded for long time spans, i.e., up to $C-1$ frames.
Given the graph formulation we introduced in Sec. \ref{background}, this is only possible if we consider edges in our graph spanning across all possible time distances. Doing so naively has two main drawbacks. First, it is prohibitively expensive for long sequences, as it requires either considering a quadratic number of edges or using sophisticated pruning techniques. 
Secondly, it  implies that most edges in the graph will represent incorrect hypotheses. To prove that, observe that each object detection can, at most, be incident to two correct edges (one in the past, and one in the future). Therefore, in a clip with $n$ object detections, the correct number of hypotheses is at most $2n$, while the number of edges can grow up to  $n^2$. The ratio  $o\left ( \frac{1}{n}\right)$ will vanish for sequences consisting of thousands of detections and will cause a severe label imbalance for learning methods operating on those edges.

%which will yield a severe edge label imbalance for %learning methods. \gui{Maaybe separate paragraph explaining this:} 

\PAR{Building a hierarchical clip partition}. Motivated by the aforementioned limitations, we  propose a hierarchy of smaller graphs that operate over long video clips instead of a single large monolithic graph. 
% Intuitve explanation
Our hierarchy is based on a recursive partition of the clip into non-overlapping time windows or smaller clips. We illustrate our construction in Figure \ref{fig:hierarchy}. At each consecutive hierarchy level we only consider edges among tracklets contained in small windows of consecutive frames, which ensures that our tracklet length is relatively uniform at every level. %\TODO{maybe argue about why disjoint intervals is good?} 
After each consecutive level in our hierarchy, we merge tracklets that are close in time into longer ones. These longer tracklets then become our new set of nodes to be associated at the following hierarchy level. %, and therefore the overall number of nodes, i.e., tracklets, decreases after each step. 
By recursively merging tracklets in nearby frames, we progressively reduce the number of nodes after each hierarchy level. 
Therefore, at each consecutive level of our hierarchy, we can consider edges spanning across longer timespans without neither prohibitively increasing the edge count, nor incurring a severe label imbalance. %\orc{grammar check: without neither is not double negation right? Is this grammatically correct. (grammarly didn't find an error but just checking.) }

\begin{figure}
\centering
\includegraphics[width=\columnwidth]{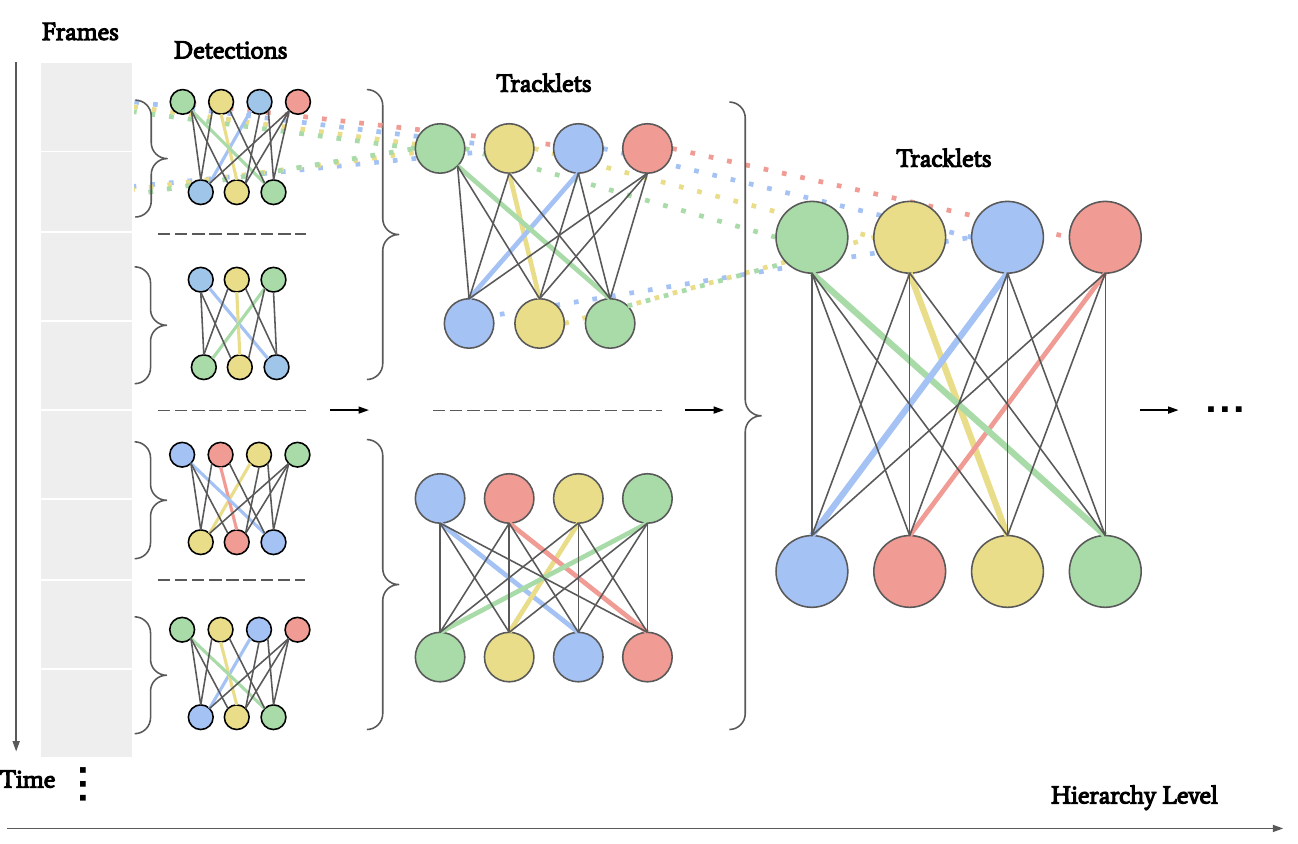}
\caption{ Our hierarchy is based on recursive partitioning of the video clip and we only allow edges within these partitions. After each level, we merge tracklets belonging to the same identity and consider edges spanning across longer timespans. }
\label{fig:hierarchy}
\end{figure}

%\subsection{Learning to track on a hierarchy of graphs} 
\subsection{Learning a unified hierarchical tracker} \label{features}
\PAR{Overview}. In the previous section, we presented a hierarchical graph-based framework to recursively merge short tracklets into longer ones. We use a message-passing GNN to process graphs in our hierarchy and learn to merge tracklets. We refer to each hierarchy level of our model as a \textit{\blockname}. A core feature of our method is that each \blockname uses the same architecture, and has access to the same feature sources. This is in contrast to previous work that engineered different solutions for different levels ~\cite{mpntrack, lift, aplift, lpc}. Therefore, instead of making assumptions on the cues needed to perform association at each timespan, we let \blocknameplural at each hierarchy \textit{learn} these from data.

\PAR{\blocknameplural.} An illustration of a \blockname is shown in Figure \ref{fig:pipeline}. Inspired by \cite{mpntrack}, the main idea behind them is: (i) consider a graph with tracklets as nodes as defined in Section~\ref{section:graph_hierarchy}, (ii) propagate node and edge embeddings across the graph via message-passing, and (iii) perform edge classification to merge nodes, \ie, tracklets, into longer tracklets. 
%
%More specifically, for each graph $G^l=(V^l, E^l)$ at level $l$ in our hierarchy (we omit time-interval indices for clarity), we consider embeddings $h^{(0)}_v\in \mathbb{R}^{d_V}$ and $h^{(0)}_{(u, w)}\in \mathbb{R}^{d_E}$ for every node $v\in V^l$ and edge $(u, w)\in E^l$, with $d_V$ and $d_E$ being their respective dimension. 
%
More specifically, for each graph $G^l=(V^l, E^l)$ at level $l$ in our hierarchy, we consider embeddings $h^{(0)}_v\in \mathbb{R}^{d_V}$ and $h^{(0)}_{(u, w)}\in \mathbb{R}^{d_E}$ for every node $v\in V^l$ and edge $(u, w)\in E^l$, with $d_V$ and $d_E$ being their respective dimension. 
Node embeddings are zero-initialized, and edge embeddings are learned from a set of initial association features.
%Node embeddings encode track-level appearance, and edge embeddings encode positional, appearance, and motion pairwise association cues. Their exact computation is detailed in section \ref{features}. 
%
The goal of message-passing is to then propagate embeddings across the graph, hereby encoding higher-order contextual information in them. To do so, we follow the time-aware neural message-passing framework of \cite{mpntrack} for $s=1, \dots, S$ steps, yielding embeddings $h^{(1)}_{(u, v)}, \dots, h^{(S)}_{(u, v)}$. 
After the last step, edge embeddings are fed to a binary classifier to obtain a score representing whether they represent a correct hypothesis or not, $y_{(u, v)}^{\text{pred}}= \mlp_{\text{class}}(h^{(S)}_{(u, v)})$. After that, predicted edge scores are rounded to obtain binary decisions using a linear program. As an end result, we obtain a longer-spanning set of tracklets from the initial ones in $V^l$. Further details on our GNN and rounding scheme are provided in the supplementary material.

\PAR{Edge association cues}. %\orc{I'd still suggest using smth like this to highlight "tracklet features are nicer than detection features". I know it is not the main selling point now but why not highlight it while we can? Ignore this if you disagree: Nodes upon which our graph neural networks operate represent tracklets. This is in contrast to previous graph-based work that used instead single detections \cite{mpntrack, lift, aplift}. We exploit this fact to obtain additional, more robust association cues that we use within the edge embeddings.} 
We compute input edge embeddings to the GNN in each \blockname by feeding an initial vector of concatenated pairwise association features to a light-weight multi-layer perception $\text{MLP}_{\text{edge}}$. %\gui{level index?}\orc{imo not needed, let's keep it simple}. 
The initial vector computation is a generalization and extension of \cite{mpntrack} to tracklets, and based on spatial 
 and motion-based proximity, time distance, and reID embedding-based appearance similarity between nodes. For appearance, we consider the average embedding vector over all detections in the tracklet which is more robust than that of a single detection. For spatial, size, and time proximity, we use the closest detections in time for each pair of nodes. Explicitly working with tracklets allows us to utilize motion, a strong cue that ~\cite{mpntrack} does not exploit. %We add a motion component to initial edge embeddings by computing the Generalized IoU \cite{rezatofighi2019generalized} of the forward and backward propagated positions by a linear motion model for every pair of tracks. For further details, we refer to Sec. 2 in the supplementary material.
 Given tracklets $T_u$ and $T_v$, we estimate their pixelwise velocities as $v_u$ and $v_v$, respectively. We then forward propagate $u$'s last position and backward propagate $v$'s first position until their middle time point $t^{\text{mid}} \eqdef (t_{v_1} - t_{u_{n_u}}) / 2$. %, assuming $t_{v_1} > t_{u_{n_u}}$. 
 Formally, we compute $pos_{u\rightarrow v}^{\text{fwrd}} \eqdef b_{u_{n_u}} + t^{\text{mid}}v_u$ and $pos_{v\rightarrow u}^{\text{bwrd}} \eqdef b_{v_1} - t^{\text{mid}}v_v$, to obtain the edge feature $GIoU(pos_{u\rightarrow v}^{\text{fwrd}}, pos_{v\rightarrow u}^{\text{bwrd}})$, where $GIoU$ is the Generalized Intersection over Union score~\cite{rezatofighi2019generalized}. For further details, we refer to the supplementary material.

\PAR{Weight sharing}. \blocknameplural use the same GNN architecture at every hierarchy level. We observe that we can \textit{share parameters and learnable weights} among the GNNs used for each SUSHI block. To do so, we additionally learn a level embedding $\phi_l$ for each level $l$ in our hierarchy. Level embeddings are added to the output edge embeddings produced by $\text{MLP}_{\text{edge}}$, and allow edge embeddings to encode the specific feature differences to be expected at each hierarchy level (e.g. larger time distances or  spatial displacements at higher levels). By sharing weights among levels, we boost the number of training samples that our GNNs have, as they now can benefit from data from multiple hierarchy levels. Empirically, we observe improved performance and convergence speed, together with a reduction in overall parameter count.

\PAR{Training.} Our message-passing GNNs are trained jointly across all levels. To do so, we sequentially unfreeze levels from first to last. This ensures that tracklets used in each subsequent level in the hierarchy during training are stable and accurate enough to provide a meaningful signal. 
Specifically, starting with only the first block being unfrozen, we unfreeze each subsequent level after $M$ training iterations, with $M=500$ in our implementation. After the GNNs from all levels are unfrozen, they are trained jointly. This allows them to adapt to the particular tracklet statistics produced by their predecessors. 
For each level, we use a focal loss~\cite{focal_loss} over the edge classification scores produced and sum losses over all levels to obtain our final loss.

%. Our overall loss is the sum of the losses for all levels. %Note, there is no gradient flow among GNNs in different levels. %In contrast to hierarchical GNNs \cite{hierarchical_gnns}, \lau{dangerous how you mention HGNN as if they were applicable to the problem, make more specific that they tackle a different task.} 
%We have empirically found that backpropagating, \eg hierarchical node features among GNNs in consecutive levels slightly decreases performance, and significantly increases memory requirements. Overall, while our model is not end-to-end differentiable, it is composed of learnable modules that can jointly learn to interact with each other's outputs.

%\PAR{Edge association cues} \TODO{}

%\PAR{Tracking multiple classes} To extend our
%$\cos \left({\angle(\rho^u_{\text{avg}}, \rho^v_{\text{avg}}})\right ) = ({\rho^u_{\text{avg}} \rho^v_{\text{avg}}}) / ({\lVert\rho^u_{\text{avg}}\lVert \lVert\rho^v_{\text{avg}}\lVert})$. \TODO{Adjust the equation according to the  ReID model.}

%\gui{Maybe discussion on limitations}
%\orc{I think these are cool but low priority}

% Experiments
\section{Experiments}
\label{sec:experiments}

\subsection{Datasets and Metrics}
We conduct experiments on four public benchmarks with significantly different properties: MOT17~\cite{motchaijcv}, MOT20~\cite{mot20}, DanceTrack~\cite{dancetrack}, and BDD100K~\cite{bdd}.
%!TEX root = ../main.tex
\begin{table*}[ht]
\center
\tabcolsep=0.11cm

    \resizebox{2\columnwidth}{!}{
    \begin{tabular}{l | c | c | c | c | c c c}
     \toprule
     Method & Short ($\leq$ 25 Frames) & Long (75 Frames) &  Long++ (150 Frames) & Very Long (512 Frames) & IDF1 & HOTA & MOTA \\ [0.5ex] 
     \midrule

    IoU & IoU & \xmark & \xmark & \xmark & 60.5 & 55.4 & 64.9 \\
    Tracktor & Reg. & \xmark & \xmark & \xmark & 65.6 & 58.9 & 66.0 \\
    IoU++ & IoU & ReID + Motion & \xmark & \xmark & 65.1 & 58.4 & 65.8  \\
    % Tracktor++ & Reg. & ReID + Motion & \xmark & 72.7 & 63.1 & 66.2 \\
    Tracktor++ & Reg. & ReID + Motion & \xmark & \xmark & 69.2 & 60.7 & 66.8 \\
    MPNTrack  & Reg. & GNN & \xmark & \xmark & 71.8 &  62.8 & 67.6 \\
    % \TODO{} & Reg. & GNN(ours) & \xmark & 72.7 & -- & 67.5 \\
    % \TODO{} 3 avg& Reg. & GNN(ours) & \xmark & 72.8 & -- & 67.6 \\
    MPNTrack++ & Reg. & GNN & \xmark & \xmark & 73.2 & 62.9 & 67.7 \\
    % (maybe)\TODO{} & GNN(ours) & ReID + Motion & \xmark & \textcolor{red}{73.0} & \textcolor{red}{63.8} & \textcolor{red}{66.4} \\
    \midrule 
    % Ours & \multicolumn{2}{|c|}{Hierarchy of 2 GNNs (ours)} & \xmark & 72.8 & -- & 67.7 \\
    % Ours 3avg & \multicolumn{2}{|c|}{Hierarchy of 2 GNNs (ours)} & \xmark & 72.9 & -- & 67.7 \\
    \modelname~(Ours) - 2 Level & \multicolumn{2}{|c|}{Hierarchy of 2 GNN} & \xmark & \xmark & 73.5 & 63.8 & 67.7 \\
     \midrule
    Tracktor++ & Reg. & \multicolumn{2}{|c|}{ReID+Motion} & \xmark & 71.2 & 62.5 & 66.4 \\

    MPNTrack & Reg. & \multicolumn{2}{|c|}{GNN} & \xmark & 71.9 & 62.8 & 67.6 \\
    % \TODO{} & Reg. & \multicolumn{2}{|c|}{GNN (ours)} & 72.8 & -- & 67.9 \\
    % \TODO{} 3avg & Reg. & \multicolumn{2}{|c|}{GNN (ours)} & 73.2 & -- & 67.9 \\
    MPNTrack++ & Reg. & \multicolumn{2}{|c|}{GNN} & \xmark & 73.4 & 63.7 & 68.0 \\
    \midrule 
    % Ours & \multicolumn{3}{|c|}{Hierarchy of 2 GNNs (ours)} & 73.2 & -- & 68.2 \\
    % Ours(3 avg) & \multicolumn{3}{|c|}{Hierarchy of 2 GNNs (ours)} & 73.5 & -- & 68.3 \\
    \modelname~(Ours) - 2 Level & \multicolumn{3}{|c|}{Hierarchy of 2 GNN} & \xmark & 73.8 & 64.2 & 68.3 \\
    \modelname~(Ours) - 3 Level & \multicolumn{3}{|c|}{Hierarchy of 3 GNN} & \xmark & 74.8 & 64.9 & 68.3 \\
    \midrule
    % \modelname~(Ours) &  \multicolumn{3}{|c|}{Hierarchy of 4 GNN} & \xmark & 76.0 & 65.8 & 68.6 \\
    MPNTrack++ & Reg. & \multicolumn{3}{|c|}{GNN} & 73.7 & 64.2 & 67.8 \\
    \midrule 
    \modelname~(Ours) - 3 Level &  \multicolumn{4}{|c|}{Hierarchy of 3 GNN} & 76.2 & 65.5 & 68.8 \\
    \modelname~(Ours) &  \multicolumn{4}{|c|}{Hierarchy of 9 GNN} & 77.6 & 66.7 & 68.9 \\
     \midrule
    \end{tabular}}
\caption{Ablation study on hybrid and unified multi-level approaches. }
\label{table:ablation_multiscale}
  \end{table*}

\PAR{MOT17.} The benchmark includes diverse scenarios with varying camera viewpoints and pedestrian densities. It consists of 14 videos (7 for training,  and 7 for testing) with a total of 11235 frames.  In the public setting, object detections are provided to emphasize data association performance during evaluation, while in the private setting trackers are allowed to use their own set of detections. 

\PAR{MOT20.} The dataset focuses on extremely crowded scenes, with some videos containing over 200 pedestrians per frame on average. It contains 8 videos, a total of 13410 frames with more than 2 million object annotations. Similar to MOT17, it includes public and private settings.

\PAR{DanceTrack.} The benchmark consists of 100 group dancing videos (105855 frames). Scenes from this dataset are moderately crowded and contain persons with highly similar appearances and complex and diverse motion patterns. %\orc{Maybe mention current sota trackers struggle due to these properties and cite Dancetrack (this is from their paper.)} \gui{yeah let's do it}

\PAR{BDD100K.} The dataset focuses on egocentric vision from a car-mounted camera in highly diverse autonomous driving scenarios. Therefore, large camera motion is prominent. The MOT benchmark contains 2,000 videos with approximately 400K images and 8 object categories.

\PAR{Metrics.}  To evaluate several aspects of MOT, we report the widely accepted metrics IDF1~\cite{idf1}, MOTA~\cite{clear}, and recently introduced HOTA~\cite{hota}. While IDF1 measures identity preservation, MOTA is heavily biased towards trajectory coverage, i.e., detection quality. HOTA finds balance between both aspects, and unifies detection, association and localization performance into a single metric. Since our method focuses on data association, we treat IDF1 and HOTA as main metrics, and also report MOTA %, false positives (FP), false negatives (FN) 
and identity switches (ID Sw.) for completeness.

\subsection{Implementation Details} 
\PAR{Model architecture and training. }   \modelname consists of nine SUSHI blocks utilizing MPNTrack~\cite{mpntrack}'s GNN architecture and shared weights between hierarchy levels. %in our \blocknameplural. 
For our reID network we use a pretrained ResNet50-IBN following \cite{lpc}. %, which replaces batch normalization layers with instance-batch-normalization layers. 
%Unlike \cite{mpntrack}, which trained its model on three different datasets: Market-1501~\cite{Zheng2015ScalablePR}, CUHK03~\cite{cuhk}, and the currently retracted DukeMTC~\cite{Ristani2016PerformanceMA}, we only pretrain our reid model on MSMT17~\cite{Wei2018PersonTG}, an extension Market-1501. Unlike previous work \cite{lpc, Tang_2017_CVPR}, 
We do not fine-tune our reID model on tracking videos, and simply freeze it during training. GNNs across hierarchical levels are trained jointly with a learning rate of $3\cdot10^{-4}$ and a weight decay of $10^{-4}$ in batches of 8 clips for 250 epochs. We use focal loss with $\gamma = 1$ %\gui{I think there's some other parameters involved} 
and the Adam optimizer~\cite{adam}. %with $\beta_1 = 0.9$ $\beta_2 = 0.999$ for all datasets.

\PAR{Hierarchy construction and inference.} Though our model is capable of processing sequences of arbitrary length, we construct hierarchies spanning a maximum temporal edge distance of 512 frames (approx. 17-36 seconds on MOT17). We experimentally found that this covers the majority of occlusions and long-term association cases across multiple datasets, and is a significant improvement over previous works, which processed clips of up to 15~\cite{mpntrack}, 32~\cite{zhou2022global}, and 60 frames~\cite{lift, aplift}.
%\orc{This is convincing right?} \gui{I think it's ok, maybe we gotta revisit it bc I might add a discussion section to talk about this}. %We restrict nodes to have edge connections within a time window size $w_l$ of 5, 25, 75 and 150 frames at each hierarchy level. 
Our hierarchy consists of nine levels and we process $2^{l}$ frames in each hierarchy level $l$. For each graph, we connect each node to its top 10 nearest neighbors based on geometry, appearance, and motion similarity. We process entire videos by feeding overlapping clips of 512 frames to our method in a sliding window fashion. We then merge per-clip tracks into trajectories of arbitrary length with a simple stitching scheme, following \cite{mpntrack}. %\orc{other methods doing similar stuff (MPNTrack, I remember at least one other method))}. 
During inference, we fill trajectory gaps by linear interpolation. 

\PAR{Runtime.} Inference runs at a competitive runtime of 21fps on MOT17 with given detections.

%Inference runs at a competitive runtime of 24fps on MOT17 and 10fps on the heavily crowded sequences of MOT20, with given detections.  %\TODO{maybe bdd, dancetrack.} %\orc{maybe:  We haven't calculated these numbers for BDD and DanceTrack though.} %\gui{let's have runtime numbers, definitely!!} 
%For further implementation details, we refer to the supplementary material.

\PAR{Object Detections.} For DanceTrack and BDD, as well as the private settings of MOT17 and MOT20, we obtain object detections from a YOLOX detector~\cite{ge2021yolox} trained following~\cite{bytetrack}. In the public setting of MOT17 and MOT20, for a fair comparison to the state-of-the-art, we use the public detections provided by MOTChallenge and preprocess them with~\cite{tracktor} %as in~\cite{aplift}. Preprocessing public detections is a widely used strategy in published methods~\cite{mpntrack, lift,  lpc, gmt}.
following~\cite{mpntrack, lift,  lpc, gmt}.

\subsection{Ablation Studies}
We conduct the ablation experiments on the MOT17 train set by performing 3-fold cross-validation following~\cite{mpntrack}. Our
ablation studies analyze two main aspects of our method: 
%\TODO{Refer to our unified method and how this is good. (See Table~\ref{table:ablation_multiscale})}, 
(i) the advantages of using a unified method across different levels of our hierarchy for data association, and
%(i) the advantages of using a unified method across  levels for data association, and
(ii) our model's ability to scale to long video clips.%, and how it leads improved identity preservation 

%and (iii) \TODO{Refer to tracklet-level features. Highlight we analyze how they influence tracking based on different hierarchy levels corresponding to time-scales. (See Figure~\ref{fig:ablation_features})}

% \begin{figure}
% \centering
% \includegraphics[width=\columnwidth]{figures/old_ablation_oracle.png}
% \caption{\TODO{Update the figure and include memory information and maybe topk}}
% \label{fig:ablation_oracle}
% \end{figure}

% \begin{figure}
% \centering
% \includegraphics[width=\columnwidth]{figures/old_ablation_depth_scores_idf1.png}
% \caption{\TODO{Update the figure}}
% \label{fig:ablation_hierarchy}
% \end{figure}

% \begin{figure}
% \centering
% \includegraphics[width=\columnwidth]{figures/old_ablation_features_idf1.png}
% \caption{\TODO{Update the figure and possibly extend it for DanceTrack}}
% \label{fig:ablation_features}
% \end{figure}

\PAR{Hybrid vs unified multi-level approaches}. %\orc{Change par name}} 
In Table \ref{table:ablation_multiscale} we analyze the performance differences between hybrid multi-level approaches and our unified GNN-based approach. We consider four association time horizons 25, 75, 150 and 512 frames, and analyze the strengths of each approach at each time horizon. We initially consider two local baselines: a framewise hungarian matching tracker (IoU), and a superior regression-based tracker (Tracktor)\cite{tracktor}. 
As additional baselines, we add long-term association levels using a combination of reID and motion cues (Iou++, Tracktor++). We observe that single-level GNNs show stronger performance at this second level (MPNTrack, MPNTrack++)\footnote{For MPNTrack++, we enhance MPNTrack  with our motion-based edge  feature. }%\footnote{We extend MPNTrack's GNN architecture for \textit{tracklets} instead of \textit{single detections} by extending the definitions of edge features to tracklets, deriving a linear motion-based edge feature, using an improved tracklet-based pruning, and introducing learnable edge embeddings for each level of our hierarchy to enable weight-sharing, and we denote it as $\text{GNN}^{\dagger}$.} 
%\orc{})
, and using GNNs at both levels achieves the best performance gains at up to 75 frames. 
%The difference between MPNTrack and MPNTrack++ is the use of motion features within initial edge embeddings, which were not exploited in MPNTrack\cite{mpntrack} and we denote as $\text{GNN}^{\dagger}$. \TODO{Update this text}}
%
This shows that the benefits of being a unified method are also applicable at moderate timespans of up to 75 frames. In the following three rows, we attempt to naively extend hybrid approaches to perform association over longer time horizons of up to 150 frames, and observe no significant improvements, due to the increased difficulty of these scenarios. In contrast, our three-level hierarchy benefits from tracking over increased timespans. 
Finally, we investigate very long time horizons in the penultimate three rows and observe that MPNTrack++ again fails to show significant improvements while our nine-level hierarchy leads to an even better tracking performance with 77.6 IDF1. 
Next, we further investigate the need for hierarchy levels.
%In the following paragraphs, we investigate the reasons behind the need for additional hierarchy levels. %, fulAs we show in the ablations below, to benefit from considering these scenarios, we require more than two hierarchy levels. %\lau{We cant end like this, why 4 levels?}
%See Table \ref{table:ablation_multiscale}. \textcolor{red}{DUMMY DUMMY DUMMY DUMMY DUMMY DUMMY DUMMY DUMMY DUMMY DUMMY DUMMY DUMMY DUMMY DUMMY DUMMY DUMMY DUMMY DUMMY DUMMY DUMMY DUMMY DUMMY DUMMY DUMMY DUMMY DUMMY DUMMY DUMMY DUMMY DUMMY DUMMY DUMMY DUMMY DUMMY DUMMY DUMMY DUMMY DUMMY DUMMY DUMMY DUMMY DUMMY DUMMY DUMMY DUMMY DUMMY DUMMY DUMMY DUMMY DUMMY DUMMY DUMMY DUMMY DUMMY DUMMY DUMMY DUMMY DUMMY DUMMY DUMMY DUMMY DUMMY DUMMY DUMMY DUMMY DUMMY DUMMY DUMMY DUMMY DUMMY DUMMY DUMMY DUMMY DUMMY DUMMY DUMMY DUMMY DUMMY DUMMY DUMMY DUMMY DUMMY DUMMY DUMMY}

\begin{figure*}
     \centering
     \begin{subfigure}[b]{0.33\textwidth}
         \centering
         \raisebox{0.15cm}{\includegraphics[width=\textwidth]{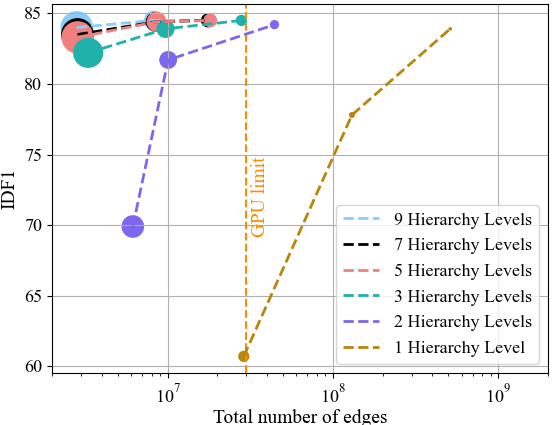}}
         \caption{Potential of hierarchies over monolithic graphs.}
         \label{fig:ablation_oracle}
     \end{subfigure}
     \hfill
     \begin{subfigure}[b]{0.36\textwidth}
         \centering
         
         \raisebox{0.12cm}{\includegraphics[width=\textwidth]{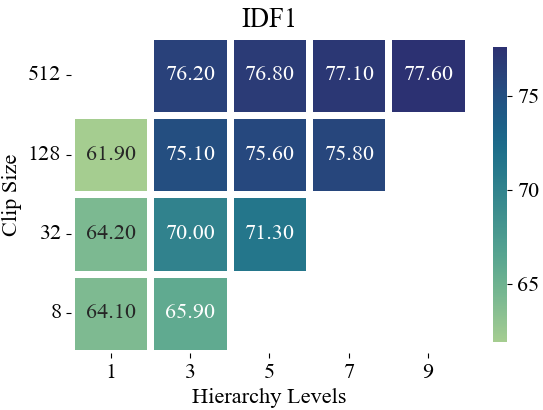}}
         \caption{Effect of clip length and hierarchy levels.}
         \label{fig:ablation_hierarchy}
     \end{subfigure}
     \hfill
     \begin{subfigure}[b]{0.30\textwidth}
         \centering
         
         \raisebox{0.25cm}{\includegraphics[width=\textwidth]{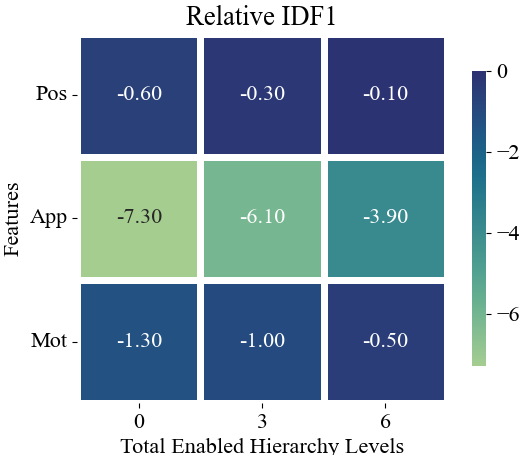}}
         \caption{Importance of features at different levels.}
         \label{fig:ablation_features}
     \end{subfigure}
        \caption{Ablation studies on scalability of our approach and association cues. }
        \label{fig:three graphs}
\end{figure*}

%\PAR{Scalability - Effect of our hierarchy on label imbalance, edge count and memory efficiency. \TODO{title change}} 
\PAR{The potential of hierarchies over monolithic graphs.} As explained in Sec.~\ref{section:graph_hierarchy}, our graph hierarchy allows us to cover large time spans with a significantly reduced number of edges, while mitigating the label imbalance between correct and incorrect trajectory hypotheses. To quantify this phenomenon, we compute the %upper bound achievable 
oracle IDF1 score, i.e., assuming perfect edge classification, of models consisting of a varying number of hierarchical levels over video clips of 512 frames. In Fig. \ref{fig:ablation_oracle}, we report both the overall number of edges needed to achieve each score, and the corresponding ratio of correct edge hypotheses, i.e., labels, %over total number of edges for all graphs in the hierarchy, 
encoded as the circle area. %A larger ratio of positive edges, hence a larger circle refers to a more balanced learning setting, which is easier to operate on for graph methods. 
For each number of levels (color-coded) each of its data points is obtained by changing the number of nearest neighbors of nodes.
%increasing the overall edge count at each graph. 
%(via increasing our KNN pruning parameter). 
We observe a clear trend: hierarchical graphs achieve significantly higher scores for a fixed memory budget. At the same time, they do so with a significantly less severe label imbalance, i.e., a larger circle area, which indicates that learning on them will be significantly easier. It is important to note that the benefits of our hierarchy saturates around nine hierarchy levels, and therefore we use this value in our model. Overall, Fig. \ref{fig:ablation_oracle}, highlights the main two reasons behind the scalability of our hierarchy.

%

%\PAR{Scalability - Effect of our hierarchy on tracking performance. \TODO{title change}} \TODO{Modify the text according to the new figure with new numbers. Most likely not much will change.}. 
%!TEX root = ../main.tex
\begin{table}[h]
\center
\tabcolsep=0.11cm

    \resizebox{\columnwidth}{!}{
    \begin{tabular}{l l c c c c c c c c c}
     \toprule
     Method & Det Ref. & IDF1 $\uparrow$ & HOTA $\uparrow$ & MOTA $\uparrow$ & ID Sw. $\downarrow$ \\ [0.5ex]

     \midrule
     \multicolumn{6}{c}{MOT17 - Public} \\
     \midrule

    % Tracktor~\cite{tracktor} & Tracktor & 55.1 & 44.8 & 56.3 & 1987 & 1.5 \\
    % MPNTrack~\cite{mpntrack} & Tracktor & 61.7 & 58.8 & 49.0 & 1185 & 35.0 \\
    % Lif\_T~\cite{lift} & Tracktor & 65.6 & 60.5 & 51.3 & 1189 & 0.5 \\
    % ApLift~\cite{aplift} & Tracktor & 65.6 & 60.5 & 51.1 & 1709 & 1.8 \\
    % GMT~\cite{gmt} & Tracktor & 65.9 & 60.2 & 51.2 & 1675 & -- \\
    % LPC\_MOT~\cite{lpc} & Tracktor & 66.8 & 59.0 & 51.5 & 1122 & 4.8 \\

    Tracktor~\cite{tracktor} & Tracktor & 55.1 & 44.8 & 56.3 & 1987 \\
    LPT~\cite{li2022learning} & Tracktor & 57.7 & -- & 57.3 & 1424 \\
    MPNTrack~\cite{mpntrack} & Tracktor & 61.7 & 49.0 & 58.8 & 1185 \\
    Lif\_T~\cite{lift} & Tracktor & 65.6 & 51.3 & 60.5 &  1189 \\
    ApLift~\cite{aplift} & Tracktor & 65.6 & 51.1  & 60.5 &  1709 \\
    GMT~\cite{gmt} & Tracktor & 65.9 & 51.2 & 60.2  & 1675 \\
    LPC\_MOT~\cite{lpc} & Tracktor & 66.8 & 51.5 & 59.0 & 1122 \\

    % GMT~\cite{gmt} & CenterTrack & 68.7 & 65.0 & 54.0 & 18213 & 177058 & 2200 \\
    % \textbf{\modelname (Ours)} & Tracktor & 67.1 & 61.5 & 52.3 & -- & --  & -- \\
    \midrule
    %\textbf{\modelname (Ours)} & Tracktor & \textbf{68.6} & \textbf{53.0} & \textbf{62.2} & \textbf{1062} \\
    \textbf{\modelname (Ours)} & Tracktor & \textbf{71.5} & \textbf{54.6} & \textbf{62.0} & \textbf{1041} \\
    %\textbf{\modelname (Ours)} & ByteTrack & xx & xx & xx & xx & xx & xx & xx \\

     \midrule
     \multicolumn{6}{c}{MOT17 - Private} \\
     \midrule
     % TraDeS~\cite{Wu2021TraDeS} & \xmark & 63.9 & 52.7 & 69.1 & 3555 & 66.9 \\
     % QDTrack~\cite{Pang_2021_CVPR} & \xmark & 66.3 & 53.9 &68.7 & 3378 & 20.3 \\
     % TrackFormer~\cite{trackformer} & \xmark & 68.0 & 74.1 & 57.3 & 2829 & 5.7 \\
     % FairMOT~\cite{zhang2021fairmot} & \xmark & 72.3 & 73.7 & 59.3 & 3303 & 25.9 \\
     % GRTU~\cite{Wang_2021_ICCV} & \xmark & 75.0 & 74.9 & 62.0 & 1812 & 3.6 \\
     % TLR~\cite{wang2021multiple} & \xmark & 73.6 & 76.5 & 60.7 & 3369 & 15.6 \\
     % ByteTrack~\cite{bytetrack} & \xmark & 77.3 & 80.3 & 63.1 & 2196 & 29.6 \\

    % TraDeS~\cite{Wu2021TraDeS} & \xmark & 63.9 & 69.1  & 52.7 & 3555  \\
    %  QDTrack~\cite{Pang_2021_CVPR} & \xmark & 66.3 & 68.7 & 53.9 & 3378  \\
    %  TrackFormer~\cite{trackformer} & \xmark & 68.0 & 74.1 & 57.3 & 2829  \\
    %  FairMOT~\cite{zhang2021fairmot} & \xmark & 72.3 & 73.7 & 59.3 & 3303  \\
    %  GRTU~\cite{Wang_2021_ICCV} & \xmark & 75.0 & 74.9 & 62.0 & 1812  \\
    %  TLR~\cite{wang2021multiple} & \xmark & 73.6 & 76.5 & 60.7 & 3369 \\
    %  ByteTrack~\cite{bytetrack} & \xmark & 77.3 & 80.3 & 63.1 & 2196 \\

     %TraDeS~\cite{Wu2021TraDeS} & \xmark & 63.9 & 52.7 & 69.1   & 3555  \\
     QDTrack~\cite{Pang_2021_CVPR} & \xmark & 66.3  & 53.9 & 68.7 & 3378  \\
     TrackFormer~\cite{trackformer} & \xmark & 68.0 & 57.3 & 74.1  & 2829  \\
     MOTR~\cite{zeng2022motr} & \xmark & 68.6 & 57.8 & 73.4 & 2439 \\
     PermaTrack~\cite{Tokmakov_2021_ICCV} & \xmark & 68.9 & 55.5 & 73.8 & 3699 \\
     MeMOT~\cite{cai2022memot} & \xmark & 69.0 & 56.9 & 72.5 & 2724 \\
     GTR~\cite{zhou2022global} & \xmark & 71.5 & 59.1 & 75.3 & 2859 \\
     FairMOT~\cite{zhang2021fairmot} & \xmark & 72.3 & 59.3 & 73.7  & 3303  \\
     GRTU~\cite{Wang_2021_ICCV} & \xmark & 75.0 & 62.0 & 74.9  & 1812  \\
     CorrTracker~\cite{wang2021multiple} & \xmark & 73.6 & 60.7 & 76.5 & 3369 \\
     Unicorn~\cite{yan2022towards} & \xmark & 75.5 & 61.7 & 77.2 & 5379 \\
     ByteTrack$^{\dagger}$~\cite{bytetrack} & \xmark & 77.1 & 62.8 & 78.9 & 2363 \\
     \textcolor{gray}{ByteTrack~\cite{bytetrack}} & \textcolor{gray}{\xmark} & \textcolor{gray}{77.3} & \textcolor{gray}{63.1} & \textcolor{gray}{80.3} & \textcolor{gray}{2196} \\
     \midrule
     % \textbf{\modelname (Ours)} & \xmark & 78.2 & 80.4 & 64.3 & -- & -- & -- \\
     %\textbf{\modelname (Ours)} & \xmark & \textbf{80.5} & \textbf{65.2} & \textbf{80.7} & \textbf{1335} \\
     \textbf{\modelname (Ours)} & \xmark & \textbf{83.1} & \textbf{66.5} & \textbf{81.1} & \textbf{1149} \\
     
     \midrule

    \end{tabular}}

\caption{Test set results on MOT17 benchmark. Det. Ref. denotes the public detection refinement strategy. As ByteTrack (gray) uses different thresholds for test set sequences and interpolation, we also report scores by disabling these as ByteTrack$^{\dagger}$ (black).}

\label{table:mot17}
\end{table}

\PAR{Effect of clip length and hierarchy levels.} In Fig. \ref{fig:ablation_hierarchy}, we visualize the impact on identity preservation (IDF1) when we increase the number of hierarchy levels and the clip length, hence incorporating longer-term association scenarios. As explained in the previous paragraph, increasing the clip size allows the tracker to potentially bridge longer occlusions, but a naive increase in size yields a very large graph with severe label imbalance. 
This is confirmed in the first column of Fig.~\ref{fig:ablation_hierarchy}: increasing the clip length in a non-hierarchical way, i.e., using a single level, can even harm performance when going beyond moderate lengths ($\geq$ 32 frames). Conversely, given a fixed clip size, increasing the number of hierarchy levels up to nine yields monotonic improvements for all clip lengths. Overall, our hierarchical framework enables processing clip sizes of up to 512 frames, with an overall improvement of +13.4 IDF1 over a naive, i.e., single level, baseline. %non-hierarchical baseline, \ie with a single level. 

\PAR{Importance of features at different levels.} %\orc{Did we ditch having these numbers also for DanceTrack to show that we "learn" to value different features. E.g. maybe reid is less important on DanceTrack. Seems like we don't have time + space in the paper for it right? Let's talk now.} 
As explained in Sec. \ref{features}, our GNNs exploit three main feature modalities: position, appearance, and motion. In Fig. \ref{fig:ablation_features}, we show our network's ability to utilize association cues differently at each hierarchy level, that is, over different association timespans. 
%e analyze how different levels in our hierarchy utilize each feature type, and show the ability of our framework . 
To do so, we report the performance loss obtained from removing edge features from each type over individual levels. 
Starting from completely removing features from each modality while keeping other modalities intact (first column), we add features of the target type sequentially over consecutive levels starting from the lowest level. %: on the second column of row one, we add position features only to the first level, while on the third column 
 %of row one, we add the to both the first and second level, and so forth.  
 We observe that appearance has the largest impact, but it is mostly used at later hierarchy levels since it leads to -3.9 IDF1 when disabled for levels 7-9 compared to -1.2 IDF1 for levels 1-3. Motion seems to have a moderate but uniform impact across levels, and lastly, position information is only relevant for short-term association. 

\subsection{Benchmark Results}

\PAR{MOT17.} Under the public setting, our model outperforms all published work using Tracktor-based preprocessing (Table~\ref{table:mot17}). It is worth noting that all listed methods are graph-based and our approach significantly surpasses their identity preservation performance, as measured by IDF1 and HOTA. Compared to MPNTrack, we achieve an improvement of 9.8 IDF1 and 5.6 HOTA, despite using their GNN architecture in our SUSHI blocks, which highlights the importance of our graph hierarchy. Analogously, we also obtain significant improvements over hybrid graph-based methods~\cite{aplift, lift, lpc}. %Additional results are shown in the supp. material.
%We report additional improvements over methods using alternative preprocessing schemes in the supplementary material. 
%
In the private setting, we significantly improve upon all published methods. While using the same detector as ByteTrack \cite{bytetrack}, our model improves upon it by 5.8 IDF1 and 3.4 HOTA, and reduces ID switches by 50\%. %Note that ByteTrack is not an online method, as it uses linear interpolation.
%!TEX root = ../main.tex
\begin{table}[h]
\center
\tabcolsep=0.11cm

    \resizebox{\columnwidth}{!}{
    \begin{tabular}{l l c c c c c c c c c}
     \toprule
     Method & Det Ref. & IDF1 $\uparrow$ & HOTA $\uparrow$ & MOTA $\uparrow$ & ID Sw. $\downarrow$ \\ [0.5ex]

     \midrule
     \multicolumn{6}{c}{MOT20 - Public} \\
     \midrule

    % Tracktor~\cite{tracktor} & Tracktor & 52.7 & 42.1 &52.6  & 1648 & 1.2 \\
    % ApLift~\cite{aplift} & Tracktor & 56.5 & 46.6 & 58.9 & 2241 & 0.4 \\
    % MPNTrack~\cite{mpntrack} & Tracktor & 59.1 & 46.8 & 57.6 & 1210 & -- \\
    % LPC\_MOT~\cite{lpc} & Tracktor & 62.5 & 49.0 & 56.3 & 1562 & 0.7 \\

    Tracktor~\cite{tracktor} & Tracktor & 52.7 & 42.1 &52.6  & 1648 \\
    LPT~\cite{li2022learning} & Tracktor & 53.5 & -- & 57.9 & 1827 \\
    ApLift~\cite{aplift} & Tracktor & 56.5 & 46.6 & 58.9 & 2241 \\
    MPNTrack~\cite{mpntrack} & Tracktor & 59.1 & 46.8 & 57.6 & 1210 \\
    LPC\_MOT~\cite{lpc} & Tracktor & 62.5 & 49.0 & 56.3 & 1562 \\
    
    \midrule
    %\textbf{\modelname (Ours)} & Tracktor & \textbf{68.2} & \textbf{53.4} & \textbf{61.6} & \textbf{1078} \\
    \textbf{\modelname (Ours)} & Tracktor & 
    \textbf{71.6} & \textbf{55.4} & \textbf{61.6} & \textbf{1053} \\
    %\textbf{\modelname (Ours)} & ByteTrack & xx & xx & xx & xx & xx & xx & xx \\

     \midrule
     \multicolumn{6}{c}{MOT20 - Private} \\
     \midrule
     TrackFormer~\cite{trackformer} & \xmark & 65.7 & 54.7 &  68.6 & 1532 \\
     MeMOT~\cite{cai2022memot} & \xmark & 66.1 & 54.1 & 63.7 & 1938 \\
     FairMOT~\cite{zhang2021fairmot} & \xmark & 67.3 & 54.6 & 61.8  & 5243 \\
     GSDT~\cite{wang2021joint} & \xmark & 67.5 & 53.6 & 67.1 & 3131\\
     CorrTracker~\cite{wang2021multiple} & \xmark & 69.1 & -- & 65.2 & 5183 \\
     ByteTrack$^{\dagger}$~\cite{bytetrack} & \xmark & 74.5 & 60.4 &  74.2 & 925 \\
     \textcolor{gray}{ByteTrack~\cite{bytetrack}} & \textcolor{gray}{\xmark} & \textcolor{gray}{75.2} & \textcolor{gray}{61.3} &  \textcolor{gray}{\textbf{77.8}} & \textcolor{gray}{1223} \\
     \midrule
     % \textbf{\modelname (Ours)} & \xmark & 78.2 & 80.4 & 64.3 & -- & -- & -- \\
     %\textbf{\modelname (Ours)} & \xmark & \textbf{77.7} & \textbf{63.1} & 74.3 & \textbf{704} \\
     \textbf{\modelname (Ours)} & \xmark & \textbf{79.8} & \textbf{64.3} & 74.3 & \textbf{706} \\
     
     \midrule

    \end{tabular}}

\caption{Test set results on MOT20 benchmark. Det. Ref. denotes the public detection refinement strategy. As ByteTrack (gray) uses different thresholds for test set sequences and interpolation, we also report scores by disabling these as ByteTrack$^{\dagger}$ (black).}

\label{table:mot20}
\end{table}

\PAR{MOT20.} %MOT20 includes more challenging sequences than MOT17 due to their significantly increased pedestrian density. 
In the crowded and challenging scenes of MOT20, we achieve even greater improvements over previous work, which further highlights the scalability benefit of our hierarchical and unified framework (Table~\ref{table:mot20}). In the public setting, we surpass all state-of-the-art approaches, remarkably, by 9.1 IDF1 and 6.4 HOTA while, again, using the same refinement model over public detections. %\TODO{Decide what to do with ByteTrack. Are we comparing with shady or same hyperparameter.} 
In the private setting, we also advance the state-of-the-art  by 4.6 IDF1 and 3.0 HOTA, while again significantly reducing ID switches. %\TODO{Bash ByteTrack} %They use inference hyperparameters tuned for every test sequence + they use interpolation on MOT17 and MOT20 datasets even though they claim to be online.  

%!TEX root = ../main.tex
\begin{table}[h]
\center
\tabcolsep=0.11cm

    \resizebox{\columnwidth}{!}{
    \begin{tabular}{l c c c c c}
     \toprule
     Method & IDF1 $\uparrow$ & HOTA $\uparrow$ & MOTA $\uparrow$ & AssA $\uparrow$ & DetA $\uparrow$  \\ [0.5ex]

     \midrule
     \multicolumn{6}{c}{DanceTrack} \\
     \midrule

     CenterTrack~\cite{centertrack} & 35.7 & 41.8 & 86.8 & 22.6 & 78.1 \\
     FairMOT~\cite{zhang2021fairmot} & 40.8 & 39.7 & 82.2 & 23.8 & 66.7 \\
     TraDes~\cite{Wu2021TraDeS} & 41.2 & 43.3 & 86.2 & 25.4 & 74.5 \\
     GTR~\cite{zhou2022global} & 50.3 & 48.0 & 84.7 & 31.9 & 72.5 \\
     QDTrack~\cite{qdtrack} & 50.4 & 54.2 & 87.7 & 36.8 & 80.1 \\
     MOTR~\cite{zeng2022motr} &  51.5 & 54.2 & 79.7 & 40.2 & 73.5 \\
     ByteTrack~\cite{bytetrack} & 53.9 & 47.7 & \textbf{89.6} & 32.1 & 71.0  \\
     %\textbf{\modelname (Ours)} & \textbf{60.7} & \textbf{61.3} & \textbf{89.9} & \textbf{46.8} & \textbf{80.5} \\
     %\textbf{\modelname (Ours)} & \textbf{63.1} & \textbf{63.0} & \textbf{88.6} & \textbf{50.2} & \textbf{79.4} \\
     \textbf{\modelname (Ours)} &  \textbf{63.4} & \textbf{63.3} & 88.7 & \textbf{50.1} &\textbf{80.1}  \\

    \midrule

    \end{tabular}}

\caption{Test set results on DanceTrack benchmark.}

\label{table:dancetrack}
\end{table}

\PAR{DanceTrack.} In Table~\ref{table:dancetrack}, we report a remarkable improvement of 9.5 IDF1 and 9.1 HOTA over state-of-the-art. %while using the same detections as~\cite{bytetrack}.
Given the unique features of this dataset, these results highlight the versatility of our approach to utilize the right cues for different scenarios. 
This is in contrast with methods like~\cite{bytetrack} that show strong performance in HOTA at both MOT17 and MOT20, but fall behind other approaches on Dancetrack. 
Our improvements are consistent across datasets, which demonstrates the generality of \modelname.
% \begin{itemize}
%    \item We are SOTA by +6.8 IDF1 and +7.1 HOTA. 
%    \item Huge difference between ours and others: Our model generalizes to scenes with highly similar appearances and diverse motion patterns
%\end{itemize}

%!TEX root = ../main.tex
\begin{table}[h]
\center
\tabcolsep=0.11cm

    \resizebox{\columnwidth}{!}{
    \begin{tabular}{l c c c c c}
     \toprule
     Method & mIDF1 $\uparrow$ & mMOTA $\uparrow$ & IDF1 $\uparrow$ & MOTA $\uparrow$ & ID Sw. $\downarrow$ \\ [0.5ex]

     \midrule
     \multicolumn{6}{c}{BDD - Validation} \\
     \midrule

    MOTR~\cite{zeng2022motr} &  43.5& 32.0 & -- & -- & -- \\
    Yu \textit{et al.}~\cite{bdd} & 44.5 &  25.9 &  66.8 &  56.9  & 8315 \\
    QDTrack~\cite{qdtrack} & 50.8 &  36.6 & 71.5 &  63.5 & \textbf{6262} \\
    TETer~\cite{li2022tracking} & 53.3 & 39.1 & -- & -- & -- \\
    Unicorn~\cite{yan2022towards} & 54.0 & 41.2 & 71.3 & 66.6 & 10876 \\
    ByteTrack~\cite{bytetrack} &  54.8 & 45.5 & 70.4 &  \textbf{69.1} & 9140 \\
    %\textbf{\modelname (Ours)} & \textbf{57.5} & \textbf{46.0} & \textbf{75.2} & \textbf{69.1} & \textbf{7283} \\
    \textbf{\modelname (Ours)} & \textbf{58.8} & \textbf{45.8} & \textbf{75.6} & 68.4 & 8556 \\
    % \textbf{\modelname (Ours)} & 58.9 & 46.0 & 76.1 & 68.7 & 7493 \\
    \midrule
    \multicolumn{6}{c}{BDD - Test} \\
    \midrule
    Yu \textit{et al.}~\cite{bdd} & 44.7 &   26.3 &   68.2 &  58.3 &  14674 \\
    QDTrack~\cite{qdtrack} & 52.3 &  35.5 & 72.3 &  64.3 & \textbf{10790} \\
    TETer~\cite{li2022tracking} & 53.3 & 37.4 & -- & -- & -- \\
    ByteTrack~\cite{bytetrack} & 55.8 & 40.1 & 71.3 &  \textbf{69.6}  &  15466 \\
    %\textbf{\modelname (Ours)} & \textbf{58.9} & \textbf{40.8} & \textbf{75.9} & \textbf{69.7} & \textbf{12076} \\
    \textbf{\modelname (Ours)} & \textbf{60.0} & \textbf{40.2} & \textbf{76.2} & 69.2 & 13626 \\
    % \textbf{\modelname (Ours)} & -- & -- & -- & -- & -- \\

    \midrule

    \end{tabular}}

\caption{Validation and test set results on BDD MOT benchmark.}

\label{table:bdd}
\end{table}

\PAR{BDD.} We further demonstrate the versatility of our approach in Table~\ref{table:bdd} where we report results on the highly diverse BDD. Note that this dataset contains multiple classes, and we simply apply our GNNs across all of them. This naive extension of our approach already achieves significant improvements of +4.2 mIDF1 and +4.9 IDF1 over the state-of-the-art. 
Overall, these results further consolidate the generality of our approach and its ability to accurately track non-pedestrian classes.

\section{Conclusion}
\label{sec:conclusion}
We have presented SUSHI, a unified method for tracking across multiple timespans. Through our ablation studies, we have shown clear benefits from the two main features of our approach: (i) its unified nature across temporal scales, and (ii) its ability to scale to long video clips.  Moreover, our benchmark results have demonstrated our model's ability  to significantly advance state-of-the-art across highly diverse tracking scenarios, hence proving its generality. 

We expect \modelname to inspire future research by questioning the need for engineering timescale-specific solutions for tracking. Lastly, we believe \modelname makes significant progress towards tackling long-term tracking, and will highlight the potential of graph hierarchies towards this end.

\smallskip
%\PAR{Acknowledgements.} \TODO{}

% For partial funding of this project, GB would like
% to acknowledge Munich Center for Machine Learning.

\newpage
%%%%%%%%% REFERENCES
{\small
\bibliographystyle{ieee_fullname}
\bibliography{egbib}
}

\clearpage
\appendix

\textbf{{\Large Supplementary Material}}
% modified and extended by Stefan Roth (stefan.roth@NOSPAMtu-darmstadt.de)
\newline

In this supplementary material, we provide further details about our method. %and (iii) an additional discussion (Sec.~\ref{sec:offline_discussion}).

%!TEX root = ../../supplementary_material.tex
\section{Additional Details about SUSHI }\label{sec:method_supplementary} 
\subsection{Edge association cues}
As explained in Section 4.2, we feed an initial vector of concatenated pairwise association features to a light-weight multi-layer perception $\text{MLP}_{\text{edge}}$ to compute input edge embeddings in each \blockname. Specifically, we obtain these features from tracklets and embed information about time distance, reID-based appearance similarity, spatial and motion-based proximity between two nodes. 
\PAR{Time and position information.} Similarly to \cite{mpntrack}, we encode the relative position and time distance among nodes as initial edge features. We naturally extend this notion from single detections to tracks by considering the closest detections in time for each pair of tracks. Formally, given the box coordinate and timestamps of two tracklets $T_u$ and $T_v$, defined as 
$u=\{(x_i, y_i, w_i, h_i, t_i)\}_{i=u_1}^{u_{n_u}}$ and
$v=\{(x_i, y_i, w_i, h_i, t_i)\}_{i=v_1}^{v_{n_v}}$, assuming $t_{u_{n_u}}< t_{v_1}$ \ie $T_u$ ends before $T_v$ starts, we compute the following position features:
$$ \left ( \frac{2(x_{n_u} - x_{v_1})}{h_{u_{n_u}} +  h_{v_1}}), \frac{2(y_{u_{n_u}} - y_{v_1})}{h_{u_{n_u}} +  h_{v_1}}), \log{ \frac{w_{u_{n_u}}}{ w_{1}}}, \log{ \frac{h_{u_{n_u}}}{ h^v_{1}}}  \right )$$ 
and we naturally compute time difference as $t_{u_{n_u}} - t_{v_1}$.

\PAR{Appearance Representation.} For every object detection $o_i\in \mathcal{O}$ we obtain embedding $\rho_i$ representing its appearance by feeding its image patch to a pretrained convolutional network, $\rho_i=\text{CNN}_{app}(a_i)$. Intuitively, while single embeddings can be affected by motion blur or sudden illumination changes, a representation summarizing the entire set can be more robust to such phenomena. Hence, we use the euclidean distance among averaged embeddings of tracks as an appearance similarity term   $\lVert\rho^u_{\text{avg}}- \rho^v_{\text{avg}}\rVert_2$.

\PAR{Motion consistency. } Trajectories are expected to be continuous in the spatio-temporal domain. We utilize this cue  by defining an additional edge feature encoding the motion consistency of each pair of tracklets. Given two tracklets $T_u$ and $T_v$, we estimate their respective velocities in the pixel domain as $v_u$ and $v_v$, respectively. Assuming again $t_{n_u}< t_{v_1}$, we use the estimated velocities to forward propagate $u$'s last position and backward propagate $v$'s first position until their middle time point $t^{\text{mid}} \eqdef (t_{v_1} - t_{u_{n_u}}) / 2$, to minimize the prediction horizon from each track. Formally, we compute $pos_{u\rightarrow v}^{\text{fwrd}} \eqdef b_{u_{n_u}} + t^{\text{mid}}v_u$ and $pos_{v\rightarrow u}^{\text{bwrd}} \eqdef b_{v_1} - t^{\text{mid}}v_v$, to obtain the edge feature $GIoU(pos_{u\rightarrow v}^{\text{fwrd}}, pos_{v\rightarrow u}^{\text{bwrd}})$, where $GIoU$ denotes the Generalized Intersection over Union score~\cite{rezatofighi2019generalized}. We choose the $GIoU$ score over the commonly used Intersection over Union because the former still provides a meaningful signal whenever two boxes do not intersect.

As explained in the main paper SUSHI blocks use the same edge features and their GNNs share weights. The only exception is the first level as motion features are not available at this level becasue each node represents a single detection. 

\subsection{Additional implementation details}
\PAR{Clip stitching. } As explained in the main paper, \modelname operates over video clips of 512 frames. To obtain trajectories over video sequences of arbitrary length, we process videos in an overlapping sliding window fashion. More specifically, we set the overlap among windows to be 256 frames and therefore process videos into clips with corresponding frame intervals (1, 512), (257, 768), (513, 1024), and so on. To stitch trajectories in overlapping windows, we use a simple bipartite matching-based algorithm. Let $\mathcal{T}_A$ and  $\mathcal{T}_B$ represent the sets of tracks in two overlapping windows, respectively, restricted over the frame interval in which they overlap. Since all trajectories in $\mathcal{T}_A$ and $\mathcal{T}_B$ are built over the same initial set of object detections, for every pair of trajectories $T_A \in \mathcal{T}_A$ and $T_B \in \mathcal{T}_B$, we can consider their $IoU$ \ie the ratio of boxes that they share in common:
$$IoU(T_A, T_B) = \frac{ \#(T_A \cap T_B)}{\#(T_A \cup T_B)}$$
Note that whenever trajectory predictions among the two clips are \textit{consistent}, their IoU will be 1, and whenever they don't share any boxes, it will be 0. Once we have computed the IoU between each pair of trajectories  $T_A \in \mathcal{T}_A$ and $T_B \in \mathcal{T}_B$, we define their pairwise cost as:
$$
c(T_A, T_B)\eqdef \begin{cases}
1 - IoU(T_A, T_B)  \  \ \ \text{if} \  \#(T_A \cap T_B) > 0 \\  \infty  \ \  \ \text{otherwise}.
\end{cases}
$$
where the second clause prevents non-overlapping tracks from being matched. Using this formulation, we obtain the min-cost bipartite matching between $\mathcal{T}_A$ and $\mathcal{T}_B$, and assign the same identity to matched trajectories.

\PAR{Edge pruning.} As mentioned in Section 5.2 of the main paper, we define the set of edges of each graph in our hierarchy by considering for each node, its top K nearest neighbors (KNNs) according to a position, appearance and motion-based similarity measure. Making graphs sparse with KNN-based edge filtering helps to reduce the number of edges, and therefore computational burden, as well as improving the edge label imbalance. Intuitively, edges between nodes with drastically different appearance or infeasible motion can be discarded early. However, there is a tradeoff: low values of $K$ might also discard edges belonging to ground truth trajectories in case of noisy features. Notably, single monolithic graphs such as MPNTrack's~\cite{mpntrack} require high values of K to achieve good performance, while in our framework consisting of relatively smaller graphs $K = 10$ suffices, yielding significantly better label distribution. 

Choosing the right distance metric to prune edges is crucial for the overall success of this strategy. Intuitively, nodes that are close should be likely to belong to the same trajectory. While MPNTrack relied solely on the distance among appearance embeddings, we take advantage of two features of our hierarchy: i) in lower hierarchy levels, nodes within the same trajectory tend to be very close in space and time ii) in higher levels, we have motion information, which can help us determine physically unreasonable connections. To exploit these facts, for graphs in the first level of our hierarchy, we simply use the coordinate-based distance between each pair of tracks as a similarity measure. In subsequent levels, for each pair of tracklets $T_u$ and $T_v$ we define their distance for edge pruning as: $d(T_u, T_v)\eqdef \lambda d_{\text{app}}(T_u, T_v) + (1-\lambda) d_{\text{motion}}(T_u, T_v)$, where $d_{\text{app}}$ is the euclidean distance of their appearance embeddings, and $d_{\text{motion}}$ is 1 minus their GIoU score. We empirically set $\lambda = 0.05$.

\subsection{Message passing network architecture}
\PAR{Time-aware neural message passing updates} 
As explained in Section 4 of the main paper, at the core of our \blocknameplural there is a message passing GNN that, given a graph at each level of our hierarchy, takes as input its initial set of node and edge embeddings, and produces new embeddings encoding high-order contextual information, that we later use for edge classification. We now explain them in detail. Formally, for each graph $G^l=(V^l, E^l)$ at level $l$ in our hierarchy, we consider embeddings $h^{(0)}_v\in \mathbb{R}^{d_V}$ and $h^{(0)}_{(u, w)}\in \mathbb{R}^{d_E}$ for every node $v\in V^l$ and edge $(u, w)\in E^l$, with $d_V$ and $d_E$ being their respective dimension. 
For a fixed number of steps, $s=1, \dots, S$ and each node $v\in V^l$ and edge $(u, v) \in E^l$ we do:
\begin{align}
h_{(u, v)}^{(s)} &=  \mlp^l_{\text{edge}}\left( \left[ h_{u}^{(s-1)}, \bar{h}_{(u, v)}^{(s-1)}, h_{v}^{(s-1)} \right] \right) \\  
m_{u\rightarrow v}^{(s)} &=\begin{cases} 
            \mlp^l_{\text{past}}\left([h_u^{(s-1)}, \bar{h}_{(u, v)}^{(s)}, h_{v}^{(s-1)}]\right) \text{if } t_u^\text{end}<t_v^\text{start}\\[7pt]
            \mlp^l_{\text{future}} \text{ }\left([h_u^{(s-1)}, \bar{h}_{(u, v)}^{(s)}, h_{v}^{(s-1)}]\right) \text{else}.  
          \end{cases}\\
h_v^{(s)} &=  \mlp^l_{\text{node}}\left ( \left [ \sum_{u|t_u^\text{end}<t_v^\text{start}} m^{(s)}_{u\rightarrow v},\sum_{u| t_u^\text{start}<t_v^\text{end}} m^{(s)}_{    u\rightarrow v} \right ] \right ) 
\end{align}

\iffalse
\begin{align}
%&(\text{edge update}) &
&(\text{Edge Update}) &h_{(u, v)}^{(s)} &=  \mlp^l_{\text{edge}}\left( \left[ h_{u}^{(s-1)}, \bar{h}_{(u, v)}^{(s-1)}, h_{v}^{(s-1)} \right] \right) \\  
&(\text{Message Computation})            &m_{u\rightarrow v}^{(s)} &=\begin{cases} 
            \mlp^l_{\text{past}}\left([h_u^{(s-1)}, \bar{h}_{(u, v)}^{(s)}, h_{v}^{(s-1)}]\right) \text{if } \text{end}(u)<\text{start}(v),\\[7pt]
            \mlp^l_{\text{future}} \text{ }\left([h_u^{(s-1)}, \bar{h}_{(u, v)}^{(s)}, h_{v}^{(s-1)}]\right) \text{otherwise}.  
          \end{cases}\\
&(\text{Node Update})            &h_v^{(s)} &=  \mlp^l_{\text{node}}\left ( \left [ \sum_{u| \text{end}(u)<\text{start}(v)} m^{(s)}_{u\rightarrow v},\sum_{u| \text{start}(u)>\text{end}(v)} m^{(s)}_{    u\rightarrow v} \right ] \right ). 
\end{align}
\fi

% \begin{align}
%     y_{(o_i, o_j)} = \begin{cases} 1 \ \ \  \text{if} \ \ \exists T_k\in \mathcal{T}^* \ \ \text{s.t.} \ \  (o_i, o_j) \in E(T_k)  \\
% 0  \ \ \ \text{otherwise}\\
% \end{cases}
% \end{align}

%
where $\mlp^l_*$ denote multi-layer perceptrons that are shared across the entire hierarchy level $l$, $[*, *]$ denotes concatenation, $\bar{h}^{(s)}_{(u, v)}\eqdef  [h^{(s)}_{(u, v)}, h^{(0)}_{(u, v)}]$ and $t_u^\text{start}$ (resp. $t_u^\text{end}(u)$) denotes the first (resp. last) timestamp of the tracklet associated to node $u\in V^l$.
Intuitively, edges are updated by combining their incident nodes' information. Then nodes are updated by separately aggregating over embeddings from their neighboring incident edges in future and past frames separately, to account for the time directionality. These updates follow the message-passing scheme in \cite{mpntrack} and, despite relying on a set of lightweight multi-layer perceptrons, they yield embeddings enabling high-accuracy edge classification.

\PAR{Detailed MLP architectures.} Our \blocknameplural consist of the MLPs defined above for neural message passing, our edge classifier $\mlp_{\text{class}}$, and an additional MLP used to intialize edge embeddings from their initial features, denoted as $\mlp^{\text{init}}_{\text{edge}}$. All of their exact architectures are detailed in Figure \ref{model_arch}. We do not count lernable per-level embeddings due to their negligible cost. Overall, our architecture not including the ResNet50-IBN reID model, has a total of approximately $22K$ parameters, which is notably small for deep learning standards.

\begin{figure*}[t!]
        \centering
           \subfloat[reID model]{%
              \includegraphics[width=0.32\textwidth]{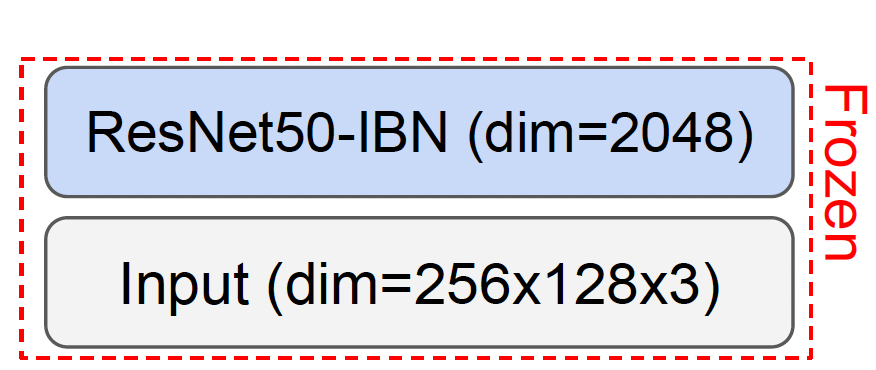}
           } 
           \subfloat[$\mlp^{\text{init}}_{\text{edge}}$]{%
              \includegraphics[width=0.32\textwidth]{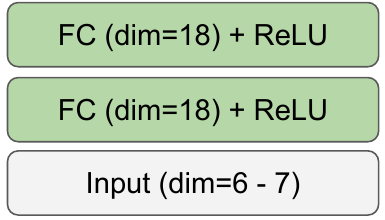}
           } 
           \subfloat[$\mlp^{\text{l}}_{\text{past}} / \mlp^{\text{l}}_{\text{future}}$]{%
              \includegraphics[width=0.32\textwidth]{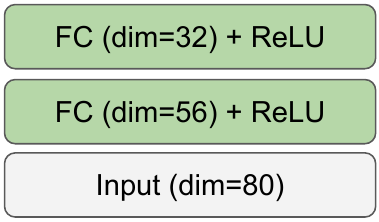}
           } 
           
           \subfloat[$ \mlp^{\text{l}}_{\text{node}}$]{%
              \includegraphics[width=0.32\textwidth]{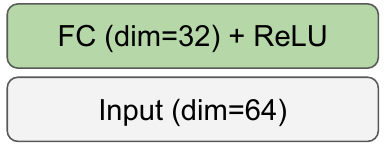}
           } 
           \subfloat[$\mlp^{\text{l}}_{\text{edge}}$]{%
              \includegraphics[width=0.32\textwidth]{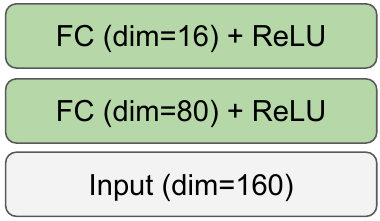}
           } 
           \subfloat[$\mlp^{\text{l}}_{\text{class}}$]{%
              \includegraphics[width=0.32\textwidth]{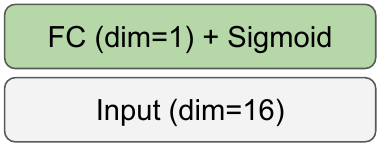}
           } 
\caption{Detailed architectures of all components of our model.} \label{model_arch}
\end{figure*}

% \subsection{Obtaining tracks via linear programming  
\subsection{Rounding edge predictions via linear programming}

As mentioned in Section 4.2, given a set of edge predictions $y^{\text{pred}}_{(u, v)\in E}$ over a graph $G=(V, E)$, we use a linear programming-based algorithm to obtain a new set of tracklets following \cite{mpntrack}. We now explain this algorithm in detail: we first look at the problem constraints and then provide the final algorithm used to enforce them.

\PAR{Flow conservation-type constraints.} Recall that edge predictions aim to approximate the set of edge labels $\{y^{\text{pred}}_{(u, v)}\}_{(u, v)\in E}$. Further recall that these labels are defined by considering the set of edges coresponding to trajectory-paths in the graph. Specifically, given a time-ordered track $T_k=\{o_{k_i}\}_{i=1}^{n_k}$ with $t_{k_i}< t_{k_{i+1}}$, we consider its corresponding path in $G$ given by its edges $E(T_k) \eqdef\{(o_{k_1}, o_{k_2}), \dots, (o_{k_{n_k - 1}}, o_{k_{n_k}}) \}$, and hence define, for each $(o_i, o_j) \in E$:
\begin{align}
    y_{(o_i, o_j)} = \begin{cases} 1 \ \ \  \text{if} \ \ \exists T_k\in \mathcal{T}^* \ \ \text{s.t.} \ \  (o_i, o_j) \in E(T_k)  \\
0  \ \ \ \text{otherwise}\\
\end{cases}
\end{align}

Now, notice that since each node (\ie object detection) can belong to at most one trajectory, edge labels need to satisfy the following constraints:
\begin{align}
 \sum_{(o_j, o_i)\in E \text{ s.t. } t_i>t_j} y_{(o_j, o_i)} &\leq 1 \ \ \ \forall o_i\in V  \label{flow_in_constr} \\
\sum_{(o_i, o_k)\in E \text{ s.t. } t_i<t_k} y_{(o_i, o_k)} &\leq 1 \ \ \ \forall o_i\in V \label{flow_out_constr}
\end{align}
Since $y$'s are binary, these constraints state that each node should have, at most, one incident edge labeled as 1 connecting it to a future (resp. past) node, and they are analogous to the conservation constraints used in network flows problems ~\cite{networkflows}.

\PAR{Projection algorithm.} The set of edge predictions, $\{y^{\text{pred}}_{(u, v)}\}_{(u, v)\in E}$ produced by our GNN already satisfies approximately 99\% of the aforementioned constraints ~\cite{mpntrack} by simply thresholding them at 0.5. In general, however,  having unsatisfied constraints makes it ambiguous to determine the trajectory of an object. In other words, if a node has two positive-labeled edges connecting it to nodes in future locations, it becomes unclear which edge should be selected to form its trajectory. To address these cases, we consider the subgraph of nodes and edges that, after thresholding, violate inequalities \ref{flow_in_constr} or \ref{flow_out_constr}, denoted as $\Bar{V}$ and $\Bar{E}$, respectively, and obtain the closest feasible binary solution $y^{\text{round}}$ to our predictions $y^{\text{pred}}$ by solving the following integer linear program:
\begin{align}
\begin{array}{ll@{}ll}
\min_{y^{\text{round}}}  &  y^{\text{round}} \left(1 - 2y^{\text{pred}}\right )      &\\
\text{subject to}& y^{\text{round}} \  \text{satisfying ineq. \ref{flow_in_constr}  and \ref{flow_out_constr} for all nodes in } \Bar{V} &\\                                               &y^{\text{round}}_{(u, v)} \in \{0,1\} \ \  \forall (u, v) \in \bar{E}
\end{array}\label{opt_projection}
\end{align}
where we index both $y^{\text{round}}$ and $y^{\text{pred}}$ indexed for all edges $(u, v)\in \Bar{E}$. Note that, since $y^\text{round}$ is binary, this objective is equivalent to minimizing the euclidean distance $\lVert y^\text{round} - y^\text{pred}\rVert_2$. Moreover, it can be shown that the constraint matrix in \ref{opt_projection} is unimodular, and hence solving the linear relaxation of the problem yields a global integer optimum \cite{berclaz2011multiple}. Overall,  Eq. \ref{opt_projection} can be very efficiently solved with off-the-shelf linear programming solvers as the graph over which it is defined has very few nodes and edges due to a high percentage of feasible edge predictions produced by our network that can be directly thresholded. %, this projection step takes a negligible portion of our model's runtime.

\end{document}